\documentclass{article}

\usepackage[final]{neurips_2024}

\usepackage[utf8]{inputenc} 
\usepackage[T1]{fontenc}    
\usepackage{url}            
\usepackage{booktabs}       
\usepackage{amsfonts}       
\usepackage{nicefrac}       
\usepackage{microtype}      
\usepackage{xcolor}         
\usepackage{colortbl}  
\usepackage{xcolor}
\usepackage{array}   
\usepackage{multirow}

\usepackage{graphicx}  
\usepackage{amsmath}   

\usepackage{enumitem}
\usepackage{balance}

\usepackage{caption}
\usepackage{subcaption}
\usepackage{subcaption}
\usepackage{pifont}
\newcommand{\cmark}{\ding{51}}%

\definecolor{citecolor}{HTML}{0071bc}
\usepackage[pagebackref=true,breaklinks=true,letterpaper=true,colorlinks,bookmarks=false,citecolor=citecolor]{hyperref}

\def\reffig{Fig.}
\def\reftab{Table}
\def\refequ{Eq.}
\def\refsec{Sec.}
\def\framework{Self-distilled Depth Refinement}

\def\sx{SDDR}
\def\sota{state-of-the-art}
\def\network{refinement network}

\def\LOSSWEIGHT{Edge-based Fusion Loss}
\def\LOSSGRAD{Edge-guided Gradient Loss}

\def\lossweight{edge-based fusion loss}
\def\lossgrad{edge-guided gradient loss}
\def\noisecon{local inconsistency noise}
\def\noiseedge{edge deformation noise}
\def\bosong{noisy Poisson fusion}
\def\BoSong{Noisy Poisson Fusion}

\title{Self-Distilled Depth Refinement \\ with \BoSong{}}

\newcommand{\authorskip}{\hspace{5mm}}

\author{%
  Jiaqi Li\textsuperscript{1,$*$} \authorskip
  Yiran Wang\textsuperscript{1,$*$} \authorskip
  Jinghong Zheng\textsuperscript{1,$*$} \authorskip \vspace{0.5mm} \\
  \textbf{Zihao Huang\textsuperscript{1} \authorskip 
  Ke Xian\textsuperscript{2} \authorskip
  Zhiguo Cao\textsuperscript{1,$\dag$} \authorskip
  Jianming Zhang\textsuperscript{3}} 
  \vspace{1.5mm} \\
  \textsuperscript{1}School of AIA, Huazhong University of Science and Technology\\
  \textsuperscript{2}School of EIC, Huazhong University of Science and Technology\\
  \textsuperscript{3}Adobe Research\\
  \vspace{1mm}
  \small{\textsuperscript{$*$}Equal contribution\hspace{0.5cm}\textsuperscript{$\dag$}Corresponding author}\\
  \texttt{\{lijiaqi\_mail,wangyiran,deepzheng,zihaohuang,kxian,zgcao\}@hust.edu.cn}\\
  \texttt{jianmzha@adobe.com}\\
  \small{\url{https://github.com/lijia7/SDDR}}
}
\DeclareMathSymbol{@}{\mathord}{letters}{"3B}

\begin{document}

\maketitle

\begin{abstract}
  Depth refinement aims to infer high-resolution depth with fine-grained edges and details, refining low-resolution results of depth estimation models. The prevailing methods adopt tile-based manners by merging numerous patches, which lacks efficiency and produces inconsistency. Besides, prior arts suffer from fuzzy depth boundaries and limited generalizability. Analyzing the fundamental reasons for these limitations, we model depth refinement as a noisy Poisson fusion problem with local inconsistency and edge deformation noises. We propose the \framework{} (\sx{}) framework to enforce robustness against the noises, which mainly consists of depth edge representation and edge-based guidance. With noisy depth predictions as input, \sx{} generates low-noise depth edge representations as pseudo-labels by coarse-to-fine self-distillation. Edge-based guidance with \lossgrad{} and \lossweight{} serves as the optimization objective equivalent to Poisson fusion. When depth maps are better refined, the labels also become more noise-free. Our model can acquire strong robustness to the noises, achieving significant improvements in accuracy, edge quality, efficiency, and generalizability on five different benchmarks. Moreover, directly training another model with edge labels produced by \sx{} brings improvements, suggesting that our method could help with training robust refinement models in future works. 

\end{abstract}

\section{Introduction}
\begin{figure*}[!t]
\centering
\includegraphics[width=1.0\textwidth,trim=0 0 0 0,clip]{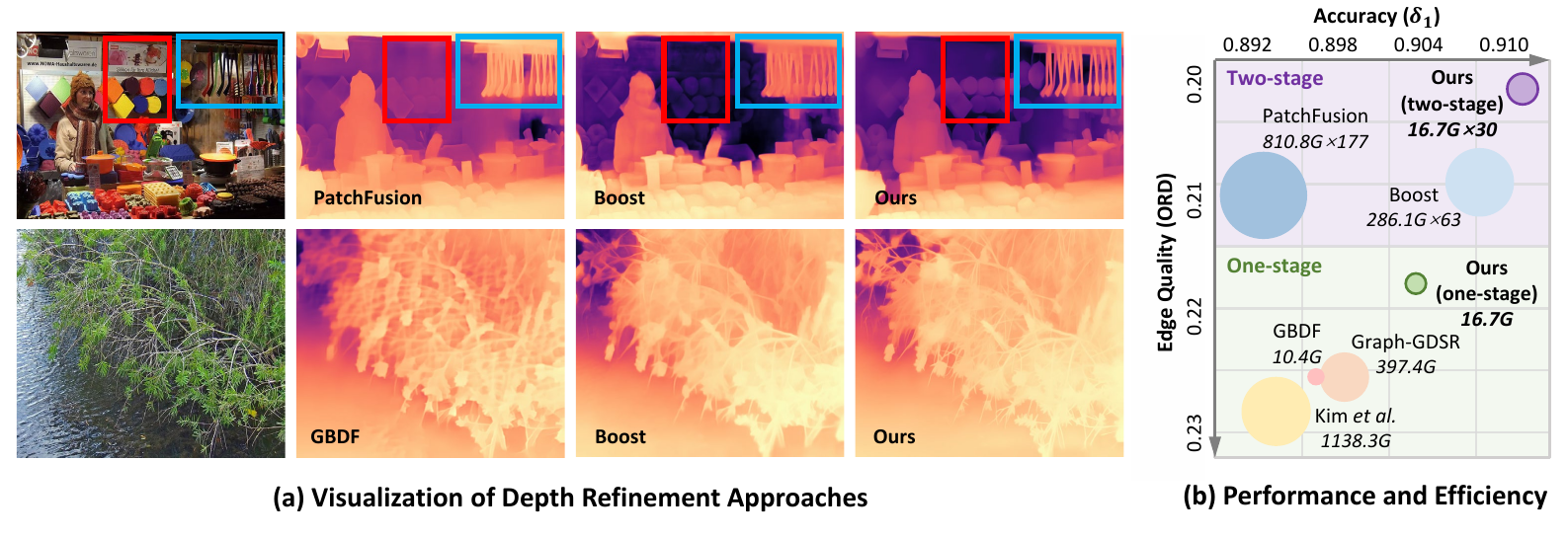}
\vspace{-18pt}
\caption{
\textbf{(a) Visual comparisons.} We model depth refinement by \bosong{} with the \noisecon{} (representing the inconsistent billboard and wall in red box) and the \noiseedge{} (indicating blurred depth edges in the blue box and second row). Better viewed when zoomed in. \textbf{(b) Performance and efficiency.} Circle area represents FLOPs. The two-stage methods~\cite{boostdepth,patchfusion} are reported by multiplying FLOPs per patch with patch numbers. \sx{} outperforms prior arts in depth accuracy ($\delta_1$), edge quality (ORD), and model efficiency (FLOPs).}
\label{fig:fig1}
\vspace{-10pt}
\end{figure*}

Depth refinement infers high-resolution depth with accurate edges and details, refining the low-resolution counterparts from depth estimation models~\cite{midas,leres,zoedepth}. With increasing demands for high resolutions in modern applications, depth refinement becomes a prerequisite for virtual reality~\cite{cvd,opengl}, bokeh rendering~\cite{bokehme,peng2022mpib}, and image generation~\cite{stabledif,controlnet}. The prevailing methods~\cite{boostdepth,patchfusion} adopt two-stage tile-based frameworks. Based on the one-stage refined depth of the whole image, they merge high-frequency details by fusing extensive patches with complex patch selection strategies. However, numerous patches lead to heavy computational costs. Besides, as in \reffig{}~\ref{fig:fig1} (a), excessive integration of local information leads to inconsistent depth structures, \textit{e.g.}, the disrupted billboard.

Apart from efficiency and consistency, depth refinement~\cite{boostdepth,layerrefine,graph-GDSR,gbdf,sun2023consistent,patchfusion} is restricted by noisy and blurred depth edges. Highly accurate depth annotations with meticulous boundaries are necessary to enforce fine-grained details. For this reason, prior arts~\cite{layerrefine,sun2023consistent,patchfusion}
only use synthetic datasets~\cite{hypersim,UnrealStereo4K,MVS-Synth,tata,irs} for the highly accurate depth values and edges. However, synthetic data falls short of the real world in realism and diversity, causing limited generalizability with blurred depth and degraded performance on in-the-wild scenarios. 
Some attempts~\cite{boostdepth,gbdf} simply adopt natural-scene datasets~\cite{nyu,nvds,kexian2020,kitti,mega} for the problem. The varying characteristics of real-world depth annotations, \textit{e.g.}, sparsity~\cite{nuscene,kitti,ddad}, inaccuracy~\cite{nyu,tum,kinect}, or blurred edges~\cite{kexian2018,nvds,wsvd,holopix50k}, make them infeasible for supervising refinement models. Thus, GBDF~\cite{gbdf} uses depth predictions~\cite{leres} as pseudo-labels, while Boost~\cite{boostdepth} leverages adversarial training~\cite{gan} as guidance. Those inaccurate pseudo-labels and guidance still lead to blurred edges as shown in \reffig{}~\ref{fig:fig1} (a). The key problem is to alleviate the noise of depth boundaries by constructing accurate edge representations and guidance.

To tackle these challenges, we dig into the underlying reasons for the limitations, instead of the straightforward merging of local details. We model depth refinement as a \bosong{} problem, decoupling depth prediction errors into two degradation components: \noisecon{} and \noiseedge{}. We use regional linear transformation perturbation as the \noisecon{} to measure inconsistent depth structures. The \noiseedge{} represents fuzzy boundaries with Gaussian blur. Experiments in \refsec{}~\ref{sec:overview} showcase that the noises can effectively depict general depth errors, serving as our basic principle to improve refinement results.

In pursuit of the robustness against the \noisecon{} and \noiseedge{}, we propose the \framework{} (\sx{}) framework, which mainly consists of depth edge representation and edge-based guidance. A refinement network is considered as the Poisson fusion operator, recovering high-resolution depth from noisy predictions of depth models~\cite{leres,midas,zoedepth}. Given the noisy input, \sx{} can generate low-noise and accurate depth edge representation as pseudo-labels through coarse-to-fine self-distillation. The edge-based guidance including \lossgrad{} and \lossweight{} is designed as the optimization objective of Poisson fusion. When depth maps are better refined, the pseudo-labels also become more noise-free. Our approach establishes accurate depth edge representations and guidance, endowing \sx{} with strong robustness to the two types of noises. Consequently, as shown in \reffig{}~\ref{fig:fig1} (b), \sx{} significantly outperforms prior arts~\cite{boostdepth,patchfusion,gbdf} in depth accuracy and edge quality. Besides, without merging numerous patches as the two-stage tile-based methods~\cite{patchfusion,boostdepth}, \sx{} achieves much higher efficiency.

We conduct extensive experiments on five benchmarks. \sx{} achieves \sota{} performance on the commonly-used Middlebury2021~\cite{middle}, Multiscopic~\cite{multiscopic}, and Hypersim~\cite{hypersim}. Meanwhile, since \sx{} can establish self-distillation with accurate depth edge representation and guidance on natural scenes, the evaluations on in-the-wild DIML~\cite{diml} and DIODE~\cite{diode} datasets showcase our superior generalizability. Analytical experiments demonstrate that these noticeable improvements essentially arise from the strong robustness to the noises. Furthermore, the precise depth edge labels produced by \sx{} can be directly used to train another model~\cite{gbdf} and yield improvements, which indicates that our method could help with training robust refinement models in future works.

In summary, our main contributions can be summarized as follows:
\begin{itemize}[leftmargin=*]
    \item[$\bullet$]We model the depth refinement task through the \bosong{} problem with \noisecon{} and \noiseedge{} as two types of depth degradation.
    \item[$\bullet$] We present the robust and efficient \framework{} (\sx{}) framework, which can generate accurate depth edge representation by the coarse-to-fine self-distillation paradigm.
    \item[$\bullet$] We design the \lossgrad{} and \lossweight{}, as the edge-based guidance to enforce the model with both consistent depth structures and meticulous depth edges. 
\end{itemize}

\section{Related Work}
\label{sec:relatedwork}
\noindent\textbf{Depth Refinement Models.} Depth refinement refines low-resolution depth from depth estimation models~\cite{midas,leres,zoedepth}, predicting high-resolution depth with fine-grained edges and details. Existing methods~\cite{gbdf,layerrefine,patchfusion,boostdepth} can be categorized into one-stage~\cite{gbdf,layerrefine} and two-stage~\cite{boostdepth,patchfusion} frameworks. One-stage methods~\cite{gbdf,layerrefine} conduct global refinement of the whole image, which could produce blurred depth edges and details. To further enhance local details, based on the globally refined results, the prevailing refinement approaches~\cite{boostdepth,patchfusion} adopt the two-stage tile-based manner by selecting and merging numerous patches. For example, Boost~\cite{boostdepth} proposes a complex patch-sampling strategy based on the gradients of input images. PatchFusion~\cite{patchfusion} improves the sampling by shifted and tidily arranged tile placement. However, the massive patches lead to low efficiency. The excessive local information produces inconsistent depth structures or even artifacts. In this paper, we propose the \framework{} (\sx{}) framework, which can predict both consistent structures and accurate details with much higher efficiency by tackling the \bosong{} problem.

\noindent\textbf{Depth Refinement Datasets.} Depth datasets with highly accurate annotations and edges are necessary for refinement models. Prior arts~\cite{patchfusion,layerrefine} utilize CG-rendered datasets~\cite{tata,irs,UnrealStereo4K,MVS-Synth,hypersim} for accurate depth, but the realism and diversity fail to match the real world. For instance, neither the UnrealStereo4K~\cite{UnrealStereo4K} nor the MVS-Synth~\cite{MVS-Synth} contain people, restricting the generalizability of refinement models. A simple idea for the problem is to leverage natural-scene data~\cite{nyu,nvds,kexian2020,kitti,mega}. However, different annotation methods lead to varying characteristics, \textit{e.g.}, sparsity of LiDAR~\cite{nuscene,kitti,ddad}, inaccurate depth of structured light~\cite{kinect,nyu,tum}, and blurred edges of stereo matching~\cite{kexian2020,nvds,wsvd}. To address the challenge, Boost~\cite{boostdepth} adopts adversarial training as guidance only with a small amount of accurately annotated real-world images. GBDF~\cite{gbdf} employs depth predictions~\cite{leres} with guided filtering~\cite{guidedfilter} as pseudo-labels. Due to the inaccurate pseudo-labels and guidance, they~\cite{gbdf,boostdepth} produce blurred edges and details. By contrast, \sx{} constructs accurate depth edge representation and edge-based guidance for self-distillation, leading to fine-grained details and strong generalizability.

\section{\sx{}: Self-Distilled Depth Refinement}
We present a detailed illustration of our \framework{} (\sx{}) framework. In \refsec{}~\ref{sec:overview}, we introduce the \bosong{} to model the depth refinement task and provide an overview to outline our approach. \sx{} mainly consists of depth edge representation and edge-based guidance, which will be described in \refsec{}~\ref{sec:represent} and \refsec{}~\ref{sec:guidance} respectively.

\subsection{\BoSong{}}
\label{sec:overview}
\begin{figure*}[!t]
\centering
\includegraphics[width=1.0\textwidth,trim=0 0 0 0,clip]{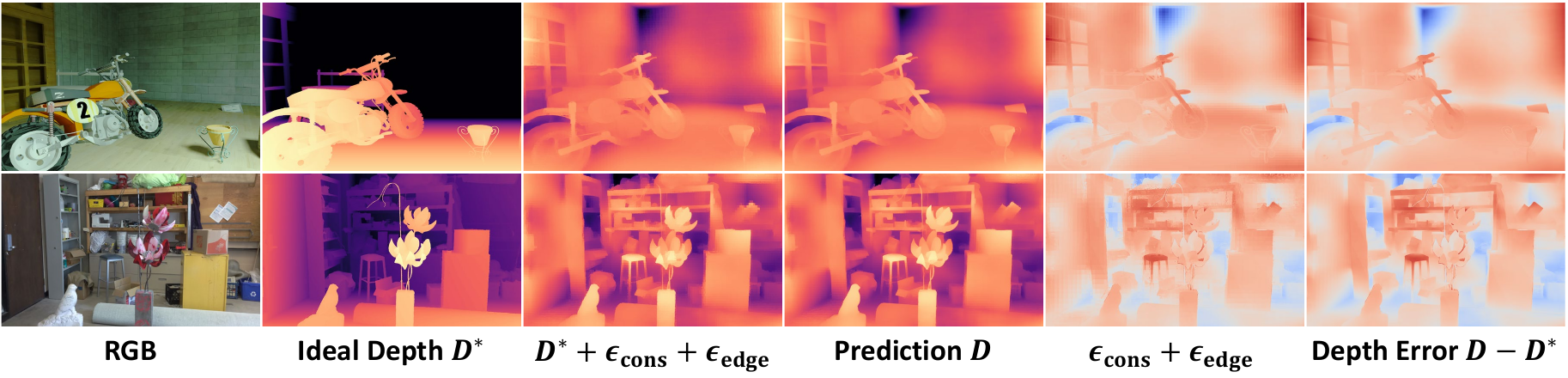}
\vspace{-15pt}
\caption{
\textbf{Depiction of depth errors.} We utilize two samples of high-quality depth maps as ideal depth $D^*$. For the predicted depth $D$, the combination of \noisecon{} $\epsilon_{\text{cons}}$ and \noiseedge{} $\epsilon_{\text{edge}}$ can approximate real depth error $D-D^*$ (the last two columns). Thus, as in the third and fourth columns, prediction $D$ can be depicted by the summation of $D^*$, $\epsilon_{\text{cons}}$, and $\epsilon_{\text{edge}}$.}
\label{fig:noise}
\vspace{-10pt}
\end{figure*}

\noindent \textbf{Problem Statement.} Based on depth maps of depth prediction models, \textit{i.e.}, depth predictor $\mathcal{N}_d$, depth refinement recovers high-resolution depth with accurate edges and details by refinement network $\mathcal{N}_r$. Some attempts in image super-resolution~\cite{ren2020real,zou2018bayesian,panagiotopoulou2012super} and multi-modal integration~\cite{li2019poisson,li2022iggm,dadp,zjrworkshop} utilize Poisson fusion to merge features and restore details. Motivated by this, we propose to model depth refinement as a \bosong{} problem. The ideal depth $D^*$ with completely accurate depth values and precise depth edges are unobtainable in real world. A general depth prediction $D$, whether produced by $\mathcal{N}_d$ or $\mathcal{N}_r$ for an input image $I$, can be expressed as a noisy approximation of $D^*$:
\begin{equation}
    D\approx D^* + \epsilon_{\text{cons}}+\epsilon_{\text{edge}}\,.
\label{eq:11111}
\end{equation}
$\epsilon_{\text{cons}}$ and $\epsilon_{\text{edge}}$ denote local inconsistency and \noiseedge{} to decouple depth prediction errors. Local inconsistency noise $\epsilon_{\text{cons}}$ represents inconsistent depth structures through regional linear transformation perturbation. Based on masked Gaussian blur, \noiseedge{} $\epsilon_{\text{edge}}$ showcases degradation and blurring of depth edges. Refer to Appendix~\ref{sec:NoiseImplementation.} for details of the noises. As in \reffig{}~\ref{fig:noise}, depth errors can be depicted by combinations of $\epsilon_{\text{cons}}$ and $\epsilon_{\text{edge}}$. Thus, considering \network{} $\mathcal{N}_r$ as a Poisson fusion operator, depth refinement can be defined as a \bosong{} problem:
\begin{equation}
\begin{gathered}
\label{eq:task}
D_0 = \mathcal{N}_r (\mathcal{N}_d(L),\mathcal{N}_d(H))\,,\\
\mathrm{s.t. } \min_{D_0,\Omega} \iint _{\Omega} \left| \nabla D_0 - \nabla D^* \right| \partial \Omega
+ \iint _{I-\Omega} \left| D_0 - D^*\right|\partial \Omega \,.
\end{gathered}
\end{equation}
The refined depth of  $\mathcal{N}_r$ is denoted as $D_0$. $\nabla$ refers to the gradient operator. Typically for depth refinement~\cite{gbdf,boostdepth,patchfusion} task, input image $I$ is resized to low-resolution $L$ and high-resolution $H$ for $\mathcal{N}_{d}$. $\Omega$ represents high-frequency areas, while $I-\Omega$ showcases low-frequency regions.

\noindent\textbf{Motivation Elaboration.} In practice, due to the inaccessibility of truly ideal depth, approximation of $D^*$ is required for training $\mathcal{N}_r$. For this reason, the optimization objective in \refequ{}~\ref{eq:task} is divided into $\Omega$ and $I-\Omega$. For the low-frequency $I-\Omega$, $D^*$ can be simply represented by the ground truth $D^*_{gt}$ of training data. However, as illustrated in \refsec{}~\ref{sec:relatedwork}, depth annotations inevitably suffer from imperfect edge quality for the high-frequency $\Omega$. It is essential to generate accurate approximations of ideal depth boundaries as training labels, which are robust to $\epsilon_{\text{cons}}$ and $\epsilon_{\text{edge}}$. Some prior arts adopts synthetic depth~\cite{UnrealStereo4K,MVS-Synth,hypersim} for higher edge quality, while leading to limited generalization capability with blurred predictions in real-world scenes. To leverage real depth data~\cite{nyu,nvds,wang2024nvds,kexian2020,kitti,mega}, GBDF~\cite{gbdf} employs depth predictions~\cite{leres} with guided filter as pseudo-labels, which still contain significant noises and result in blurred depth. Besides, optimization of $\Omega$ is also ignored. Kim \textit{et al.}~\cite{layerrefine} relies on manually annotated $\Omega$ regions as input. GBDF~\cite{gbdf,midas,dpt} omits the selection of $\Omega$ and supervises depth gradients on the whole image. Inaccurate approximations of $\nabla D^*$ and inappropriate division of $\Omega$ lead to limited robustness to \noisecon{} and \noiseedge{}.

\begin{figure*}[!t]
\centering
\includegraphics[width=1.0\textwidth,trim=0 0 0 0,clip]{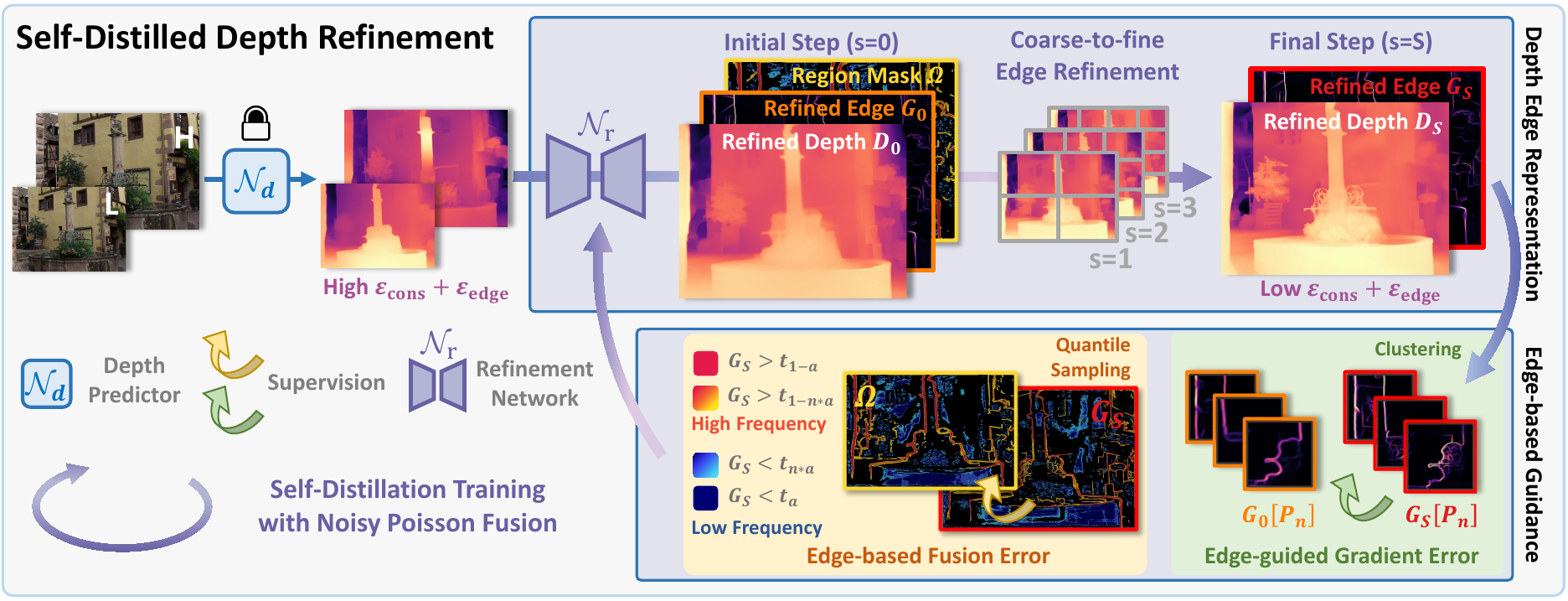}
\vspace{-15pt}
\caption{
\textbf{Overview of self-distilled depth refinement.} \sx{} consists of depth edge representation and edge-based guidance. Refinement network $\mathcal{N}_r$ produces initial refined depth $D_0$, edge representation $G_0$, and learnable soft mask $\Omega$ of high-frequency areas. The final depth edge representation $G_S$ is updated from coarse to fine as pseudo-labels. The edge-based guidance with \lossgrad{} and \lossweight{} supervises $\mathcal{N}_r$ to achieve consistent structures and fine-grained edges.}
\label{fig:pipeline}
\vspace{-10pt}
\end{figure*}

\noindent\textbf{Method Overview.} To address the challenges, as shown in \reffig{}~\ref{fig:pipeline}, we propose our \sx{} framework with two main components: depth edge representation and edge-based guidance. To achieve low-noise approximations of $\nabla D^*$, we construct the depth edge representation $G_s$ through coarse-to-fine self-distillation, where $s\in\{1,2,\cdots ,S\}$ refers to iteration numbers. The input image is divided into several windows with overlaps from coarse to fine. For instance, we denote the high-frequency area of a certain window $w$ in iteration $s$ as $\Omega_s^w$, and the refined depth of $\mathcal{N}_r$ as $D_s^w$. In this way, the self-distilled optimization of depth edge representation $G_s$ can be expressed as follows:
\begin{equation}
\begin{gathered}
D_s^w \approx D^* +\epsilon_{\text{cons}}+\epsilon_{\text{edge}} \,,\\
\min_{G_s} \sum_{w} \iint _{\Omega_s^w} \left| G_s^w - \nabla D_s^w \right| \partial \Omega_s^w \,.
\label{equation:eq3}
\end{gathered}
\end{equation}
During training, depth edge representation $G_s^w$ is further optimized based on the gradient of current refined depth $D_s^w$. The final edge representation $G_S$ of the whole image will be utilized as the pseudo-label to supervise the refinement network $\mathcal{N}_r$ after $S$ iterations. \sx{} can generate low-noise and robust edge representation, mitigating the impact of $\epsilon_{\text{cons}}$ and $\epsilon_{\text{edge}}$ (More results in Appendix~\ref{sec:app_representation}).

With $G_S$ as the training label, the next is to enforce $\mathcal{N}_r$ with robustness to the noises, achieving consistent structures and meticulous boundaries. To optimize $\mathcal{N}_r$, we propose edge-based guidance as an equivalent optimization objective to \bosong{} problem, which is presented by:
\begin{equation}
\begin{gathered}
\min_{D0,\Omega} \iint _{\Omega} \left| \nabla D_0 - G_S \right| \partial \Omega + \iint _{I-\Omega} \left| D_0 - D^*_{gt}\right|\partial \Omega \,.
\label{equation:eq4}
\end{gathered}
\end{equation}
For the second term of $I-\Omega$, we adopt depth annotations $D^*_{gt}$ as the approximation of $D^*$. For the first term, with the generated $G_S$ as pseudo-labels of $\nabla D^*$, we propose \lossgrad{} and \lossweight{} to optimize $D_0$ and $\Omega$ predicted by $\mathcal{N}_r$. The \lossgrad{} supervises the model to consistently refine depth edges with local scale and shift alignment. The \lossweight{} guides $\mathcal{N}_r$ to adaptively fuse low- and high-frequency features based on the learned soft region mask $\Omega$, achieving balanced consistency and details by quantile sampling. 

Overall, when depth maps are better refined under the edge-based guidance, the edge representation also becomes more accurate and noise-free with the carefully designed coarse-to-fine manner. The self-distillation paradigm can be naturally conducted based on the \bosong{}, enforcing our model with strong robustness against the \noisecon{} and \noiseedge{}.

\subsection{Depth Edge Representation}
\label{sec:represent}
To build the self-distilled training paradigm, the prerequisite is to construct accurate and low-noise depth edge representations as pseudo-labels. Meticulous steps are designed to generate the representations with both consistent structures and accurate details.

\noindent \textbf{Initial Depth Edge Representation.} We generate an initial depth edge representation based on the global refinement results of the whole image. For the input image $I$, we obtain the refined depth results $D_0$ from $\mathcal{N}_r$ as in \refequ{}~\ref{eq:task}.
Depth gradient $G_0=\nabla D_0$ is calculated as the initial representation. An edge-preserving filter~\cite{tomasi1998bilateral} is applied on $G_0$ to reduce noises in low-frequency area $I-\Omega$. With global information of the whole image, $G_0$ can preserve spatial structures and depth consistency. It also incorporates certain detailed information from the high-resolution input $H$. To enhance edges and details in high-frequency region $\Omega$, we conduct coarse-to-fine edge refinement in the next step.

\noindent \textbf{Coarse-to-fine Edge Refinement.} The initial $D_0$ is then refined from course to fine with $S$ iterations to generate final depth edge representation. For a specific iteration $s\in\{1,2,\cdots ,S\}$, we uniformly divide input image $I$ into ${\left(s+1\right)}^2$ windows with overlaps. We denote a certain window $w$ in iteration $s$ of the input image $I$ as $I_s^w$. The high-resolution $H_s^w$ is then fed to the depth predictor $\mathcal{N}_d$. $D_{s-1}^{w}$ represents the depth refinement results of the corresponding window $w$ in the previous iteration $s-1$. The refined depth $D_{s}^{w}$ of window $w$ in current iteration $s$ as \refequ{}~\ref{equation:eq3} can be obtained by $\mathcal{N}_d$ and $\mathcal{N}_r$:
\begin{equation}
    D_{s}^{w} = \mathcal{N}_{r}(D_{s-1}^{w},\mathcal{N}_{d}(H_{s}^{w})), 
    s \in \{1,2,\cdots ,S\}\,,
\end{equation}
After that, depth gradient $\nabla D_s^w$ is used to update the depth edge representation. The coarse-to-fine manner achieves consistent spatial structures and accurate depth details with balanced global and regional information. In the refinement process, only limited iterations and windows are needed. Thus, \sx{} achieves much higher efficiency than tile-based methods~\cite{boostdepth,patchfusion}, as shown in \refsec{}~\ref{sec:efficient}.

\begin{figure*}[!t]
\centering
\includegraphics[width=1.0\textwidth,trim=0 0 0 0,clip]{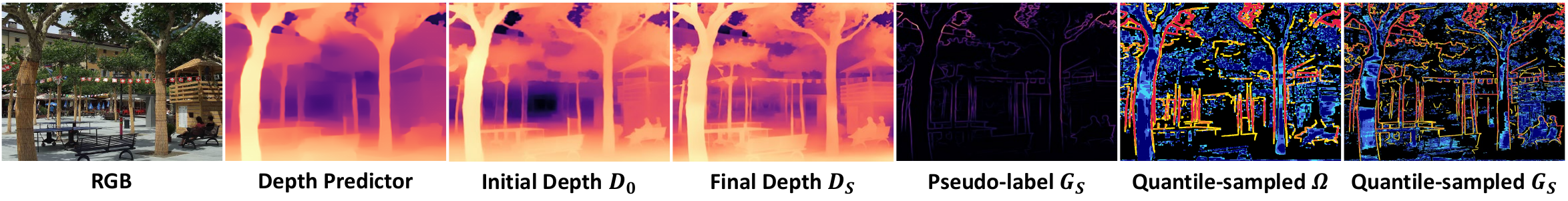}
\vspace{-15pt}
\caption{
\textbf{Visualization of intermediate results.} We visualize the results of several important steps within the \sx{} framework. The quantile sampling utilizes the same color map as in \reffig{}~\ref{fig:pipeline}.}
\label{fig:vismethod}
\vspace{-10pt}
\end{figure*}

\noindent \textbf{Scale and Shift Alignment.}
The windows are different among varied iterations. Depth results and edge labels on corresponding window $w$ of consecutive iterations could be inconsistent in depth scale and shift. Therefore, alignment is required before updating the depth edge representation:
\begin{equation}
\begin{gathered}
\label{eq:localalign}
    (\beta_1, \beta_0) = \mathop{\arg\min}\limits_{\beta_1, \beta_0}\| (\beta_1\nabla{D_{s}^{w}} + \beta_0) - G_{s-1}^w\|_2^2\,,\\
    G_s^w = \beta_1\nabla{D_{s}^{w}} + \beta_0 \,, \\
\end{gathered}
\end{equation}
where $\beta_1$ and $\beta_0$ are affine transformation coefficients as scale and shift respectively. The aligned $G_s^w$ represents the depth edge pseudo-labels for image patch $I_s^w$ generated from the refined depth $D_s^w$. At last, after $S$ iterations, we can obtain the pseudo-label $G_S$ as the final depth edge representation for self-distillation. For better understanding, we showcase visualization of $D_0$, $D_S$, and $G_S$ in \reffig{}~\ref{fig:vismethod}. 

\noindent \textbf{Robustness to Noises.} In each window, we merge high-resolution $\mathcal{N}_d(H_s^w)$ to enhance details and suppress $\epsilon_{\text{edge}}$. Meanwhile, coarse-to-fine window partitioning and scale alignment mitigate $\epsilon_{\text{cons}}$ and bring consistency. Thus, $G_S$ exhibits strong robustness to the two types of noises by self-distillation.

\subsection{Edge-based Guidance}
\label{sec:guidance}
With depth edge representation $G_S$ as pseudo-label for self-distillation, we propose the edge-based guidance including \lossgrad{} and \lossweight{} to supervise $\mathcal{N}_r$. 

\noindent\textbf{\LOSSGRAD{}.}
We aim for fine-grained depth by one-stage refinement, while the two-stage coarse-to-fine manner can further improve the results. Thus, \lossgrad{} instructs the initial $D_0$ with the accurate $G_S$. Some problems need to be tackled for this purpose. 

As $\mathcal{N}_r$ has not converged in the early training phase, $G_S$ is not sufficiently reliable with inconsistent scales and high-level noises between local areas. Therefore, we extract several non-overlapping regions $P_n, n\in\{1,2,\cdots,N_g\}$ with high gradient density by clustering~\cite{kmeans}, where $N_g$ represents the number of clustering centroids. The \lossgrad{} is only calculated inside $P_n$ with scale and shift alignment. By doing so, the model can focus on improving details in high-frequency regions and preserving depth structures in flat areas. The training process can also be more stable. The \lossgrad{} can be calculated by:
\begin{equation}
\label{eq:lossgrad}
    \mathcal{L}_{grad}=\frac{1}{N_g}\sum_{n=1}^{N_g}
    \left|\left|\left(\beta_1G_0\left[P_n\right]+\beta_0\right)-G_S\left[P_n\right]\right|\right|_1\,, 
\end{equation}
where $\beta_1$ and $\beta_0$ are the scale and shift coefficients similar to \refequ{}~\ref{eq:localalign}. We use $\left[\cdot\right]$ to depict mask fetching operations, \textit{i.e.}, extracting local area $P_n$ from $G_0$ and $G_S$. With the \lossgrad{}, \sx{} predicts refined depth with meticulous edges and consistent structures.

\noindent\textbf{\LOSSWEIGHT{}.} High-resolution feature $F_H$ extracted from $H$ brings finer details but could lead to inconsistency, while the low-resolution feature $F_L$ from $L$ can better maintain depth structures. $\mathcal{N}_r$ should primarily rely on $F_L$ for consistent spatial structures within low-frequency $I-\Omega$, while it should preferentially fuse $F_H$ for edges and details in high-frequency areas $\Omega$. The fusion of $F_L$ and $F_H$ noticeably influence the refined depth. However, prior arts~\cite{layerrefine,gbdf,boostdepth} adopt manually-annotated $\Omega$ regions as fixed masks or even omit $\Omega$ as the whole image, leading to inconsistency and blurring. To this end, we implement $\Omega$ as a learnable soft mask, with quantile sampling strategy to guide the adaptive fusion of $F_L$ and $F_H$. The fusion process is expressed by:
\begin{equation}
    F = (1-\Omega) \odot F_{L}+\Omega
    \odot F_{H}\,,
\end{equation}
where $\odot$ refers to the Hadamard product. $\Omega$ is the learnable mask ranging from zero to one. Larger values in $\Omega$ showcases higher frequency with denser edges, requiring more detailed information from the high-resolution feature $F_H$. Thus, $\Omega$ can naturally serve as the fusion weight of $F_L$ and $F_H$.

To be specific, we denote the lower quantile of $G_S$ as $t_a$, \textit{i.e.}, $P(X<t_a|X \in G_S)=a$. $\{G_S < t_{a}\}$ indicates flat areas with low gradient magnitude, while $\{G_S > t_{1 - a}\}$ represents high-frequency regions. $\Omega$ should be larger in those high-frequency areas $\{G_S > t_{1 - a}\}$ and smaller in the flat regions $\{G_S < t_{a}\}$. This suggests that $G_S$ and $\Omega$ should be synchronized with similar data distribution. Thus, if we define the lower quantile of $\Omega$ as $T_a$, \textit{i.e.}, $P(X<T_a|X \in \Omega)=a$, an arbitrary pixel $i\in \{G_S<t_{a}\}$ in flat regions should also belong to $\{\Omega<T_{a}\}$ with a lower weight for $F_H$, while the pixel $i\in \{G_S>t_{1-a}\}$ in high-frequency areas should be contained in $\{\Omega>T_{1-a}\}$ for more detailed information. The \lossweight{} can be depicted as follows:
\begin{align}
\mathcal{L}_{fusion}=\frac{1}{N_wN_p}\sum_{n=1}^{N_w}\sum_{i=1}^{N_p} \left\{
\renewcommand{\arraystretch}{1.2}
\begin{array}{ll}
\!\!\max(0,\Omega_i-T_{n*a}),   & i\in \{G_S<t_{n*a}\}\,, \\
\!\!\max(0,T_{1-n*a}-\Omega_i), & i\in \{G_S>t_{1-n*a}\}\,,\\
\end{array}
\right.
\label{equation:lossfusion}
\end{align}
where $N_p$ is the pixel number. We supervise the distribution of $\Omega$ with lower quantiles $T_{n*a}$ and $T_{1-n*a}, n\in\{1,2,\cdots,N_w\}$. Therefore, pixels with larger deviations between $G_S$ and $\Omega$ will be penalized more heavily. Taking the worst case as an example, if $i\in \{G_S<t_{N_w*a}\}$ but $i\notin \{\Omega<T_{N_w*a}\}$, the error for the pixel will be accumulated for $N_w$ times from $a$ to $N_w*a$. $\mathcal{L}_{fusion}$ enforces \sx{} with consistent structures (low $\epsilon_{\text{cons}}$ noise) in $I-\Omega$ and accurate edges (low $\epsilon_{\text{edge}}$ noise) in $\Omega$. The visualizations of quantile-sampled $G_S$ and $\Omega$ are presented in \reffig{}~\ref{fig:vismethod}.

Finally, combining $\mathcal{L}_{grad}$ and $\mathcal{L}_{fusion}$ as edge-based guidance for self-distillation, the overall loss $\mathcal{L}$ for training $\mathcal{N}_r$ is calculated as \refequ{}~\ref{eq:lossall}. $\mathcal{L}_{gt}$ supervises the discrepancy between $D_0$ and ground truth $D^*_{gt}$ with affinity-invariant loss~\cite{midas,dpt}. See Appendix~\ref{sec:A} for implementation details of \sx{}.
\begin{equation}
\begin{gathered}
\label{eq:lossall}
    \mathcal{L}=\mathcal{L}_{gt}+\lambda_1\mathcal{L}_{grad}
    +\lambda_2\mathcal{L}_{fusion}\,.
\end{gathered}
\end{equation}

\section{Experiments}
\label{sec:experi}
To prove the efficacy of \framework{} (\sx{}) framework, we conduct extensive experiments on five benchmarks~\cite{middle,multiscopic,hypersim,diml,diode} for indoor and outdoor, synthetic and real-world. 

\noindent \textbf{Experiments and Datasets.} Firstly, we follow prior arts~\cite{gbdf,boostdepth,layerrefine} to conduct zero-shot evaluations on Middlebury2021~\cite{middle}, Multiscopic~\cite{multiscopic}, and Hypersim~\cite{hypersim}. To showcase our superior generalizability, we compare different methods on DIML~\cite{diml} and DIODE~\cite{diode} with diverse natural scenes. Moreover, we prove the higher efficiency of \sx{} and undertake ablations on our specific designs. 

\noindent \textbf{Evaluation Metrics.} 
Evaluations of depth accuracy and edge quality are necessary for depth refinement models. For edge quality, we adopt
the $\text{ORD}$ and $\text{D}^\text{3}\text{R}$ metrics following Boost~\cite{boostdepth}. For depth accuracy, we adopt the widely-used $\text{REL}$ and $\delta_i\,(i=1,2,3)$. See Appendix~\ref{sec:B} for details.

\begin{table*}[t]
    \begin{center}
    \addtolength{\tabcolsep}{+1.3pt}
    \resizebox{0.985\textwidth}{!}{
    \begin{tabular}{llcccccccccccc}
    \toprule
    \multirow{2.5}{*}{Predictor} & \multirow{2.5}{*}{Method}  &
    \multicolumn{3}{c}{Middlebury2021} &
    \multicolumn{5}{c}{Multiscopic} &
    \multicolumn{3}{c}{Hypersim} \\
    \cmidrule{3-5} \cmidrule{7-9} \cmidrule{11-13}  
    &&$\delta_1\!\!\uparrow$ & $\text{REL}\!\downarrow$ & $\text{ORD}\!\downarrow$ & & 
    $\delta_1\!\!\uparrow$ & $\text{REL}\!\downarrow$ & $\text{ORD}\!\downarrow$ & & 
    $\delta_1\!\!\uparrow$ & $\text{REL}\!\downarrow$ & $\text{ORD}\!\downarrow$ \\
    \midrule
    \multirow{5}{*}{MiDaS}      & MiDaS~\cite{midas}    
    &$0.868$&$0.117$&$0.384$  &&$0.839$&$0.130$&$0.292$  &&$0.781$&$0.169$&$0.344$ \\
                                & Kim \textit{et al.}~\cite{layerrefine}   
    &$0.864$&$0.120$&$0.377$  &&$0.839$&$0.130$&$0.293$  &&$0.778$&$0.175$&$0.344$  \\
                                & Graph-GDSR~\cite{graph-GDSR}   
    &$0.865$&$0.121$&$0.380$  &&$0.839$&$0.130$&$0.292$  &&$0.781$&$0.169$&$0.345$  \\
                                & GBDF~\cite{gbdf}    
    &$0.871$&$0.115$&$0.305$  &&$0.841$&$0.129$&$0.289$  &&$0.787$&$0.168$&$0.338$  \\
                                & Ours      
    &$\textbf{0.879}$&$\textbf{0.112}$&$\textbf{0.299}$  &&$\textbf{0.852}$&$\textbf{0.122}$&$\textbf{0.267}$  &&$\textbf{0.791}$&$\textbf{0.166}$&$\textbf{0.318}$  \\

    \arrayrulecolor{black}\midrule
    \multirow{5}{*}{LeReS}      & LeReS~\cite{leres}    
    &$0.847$&$0.123$&$0.326$  &&$0.863$&$0.111$&$0.272$  &&$0.853$&$0.123$&$0.279$ \\
                                & Kim \textit{et al.}~\cite{layerrefine}   
    &$0.846$&$0.124$&$0.328$  &&$0.860$&$0.113$&$0.286$  &&$0.850$&$0.125$&$0.286$  \\
                                & Graph-GDSR~\cite{graph-GDSR}   
    &$0.847$&$0.124$&$0.327$  &&$0.862$&$0.111$&$0.273$  &&$0.852$&$0.123$&$0.281$  \\
                                & GBDF~\cite{gbdf} 
    &$0.852$&$0.122$&$0.316$  &&$0.865$&$0.110$&$0.270$  &&$0.857$&$0.121$&$0.273$  \\
                                & Ours
    &$\textbf{0.862}$&$\textbf{0.120}$&$\textbf{0.305}$  &&$\textbf{0.870}$&$\textbf{0.108}$&$\textbf{0.259}$  &&$\textbf{0.862}$&$\textbf{0.120}$&$\textbf{0.273}$  \\

    \arrayrulecolor{black}\midrule
    \multirow{5}{*}{ZoeDepth}   & ZoeDepth~\cite{zoedepth}
    &$0.900$&$0.104$&$0.225$  &&$0.896$&$0.097$&$0.205$  &&$0.927$&$0.088$&$0.198$ \\
                                & Kim \textit{et al.}~\cite{layerrefine}   
    &$0.896$&$0.107$&$0.228$  &&$0.890$&$0.099$&$0.204$  &&$0.923$&$0.091$&$0.204$  \\
                                & Graph-GDSR~\cite{graph-GDSR}  
    &$0.901$&$0.103$&$0.226$  &&$0.895$&$0.096$& $0.208$ &&$0.926$&$0.089$&$0.199$ \\
                                & GBDF~\cite{gbdf} 
    &$0.899$&$0.105$&$0.226$  &&$0.897$&$0.096$&$0.207$  &&$0.925$&$0.089$&$0.199$ \\
                                & Ours
    &$\textbf{0.905}$&$\textbf{0.100}$&$\textbf{0.218}$  &&$\textbf{0.904}$&$\textbf{0.092}$&$\textbf{0.199}$  &&$\textbf{0.930}$&$\textbf{0.086}$&$\textbf{0.191}$  \\
    \bottomrule
    \end{tabular}
    }
\end{center}
\vspace{-6pt}
\caption{\textbf{Comparisons with one-stage methods.} As prior arts~\cite{layerrefine,graph-GDSR,gbdf}, we conduct evaluations with different depth predictors~\cite{midas,leres,zoedepth}. For each predictor, we report the initial metrics and results of refinement methods. Best performances with each depth predictors~\cite{midas,leres,zoedepth} are in boldface.}
\vspace{-8pt}
\label{tab:tab1}
\end{table*}

\begin{table*}[t]
    \begin{center}
    \addtolength{\tabcolsep}{+1.3pt}
    \resizebox{0.95\textwidth}{!}{
    \begin{tabular}{llcccccccccccc}
    \toprule
    \multirow{2.5}{*}{Predictor} & \multirow{2.5}{*}{Method}  &
    \multicolumn{3}{c}{Middlebury2021} &
    \multicolumn{5}{c}{Multiscopic} &
    \multicolumn{3}{c}{Hypersim} \\
    \cmidrule{3-5} \cmidrule{7-9} \cmidrule{11-13}
    &&$\delta_1\!\!\uparrow$ & $\text{REL}\!\downarrow$ & $\text{ORD}\!\downarrow$ & & 
    $\delta_1\!\!\uparrow$ & $\text{REL}\!\downarrow$ & $\text{ORD}\!\downarrow$ & & 
    $\delta_1\!\!\uparrow$ & $\text{REL}\!\downarrow$ & $\text{ORD}\!\downarrow$ \\
    \midrule
    \multirow{3}{*}{MiDaS} & MiDaS~\cite{midas}    
    &$0.868$&$0.117$&$0.384$  &&$0.839$&$0.130$&$0.292$  &&$0.781$&$0.169$&$0.344$ \\
    &Boost~\cite{boostdepth}    
    &$0.870$&$0.118$&$0.351$  &&$0.845$&$0.126$&$0.282$  &&$0.794$&$0.161$&$0.332$  \\
    &Ours     
    &$\textbf{0.871}$&$\textbf{0.115}$&$\textbf{0.303}$  &&$\textbf{0.858}$&$\textbf{0.120}$&$\textbf{0.263}$  &&$\textbf{0.799}$&$\textbf{0.154}$&$\textbf{0.322}$  \\

    \arrayrulecolor{black}\midrule
    \multirow{3}{*}{LeReS}      & LeReS~\cite{leres}    
    &$0.847$&$0.123$&$0.326$  &&$0.863$&$0.111$&$0.272$  &&$0.853$&$0.123$&$0.279$ \\
    & Boost~\cite{boostdepth} 
    &$0.844$&$0.131$&$0.325$  &&$0.860$&$0.112$&$0.278$  &&$\textbf{0.865}$&$\textbf{0.118}$&$0.272$  \\
    & Ours
    &$\textbf{0.861}$&$\textbf{0.123}$&$\textbf{0.309}$  &&$\textbf{0.870}$&$\textbf{0.109}$&$\textbf{0.268}$  &&$0.858$&$0.123$&$\textbf{0.271}$  \\

    \arrayrulecolor{black}\midrule
    \multirow{4}{*}{ZoeDepth}   & ZoeDepth~\cite{zoedepth}
    &$0.900$&$0.104$&$0.225$  &&$0.896$&$0.097$&$0.205$  &&$0.927$&$0.088$&$0.198$ \\
    & Boost~\cite{boostdepth} 
    &$0.911$&$0.099$&$0.210$  &&$\textbf{0.910}$&$0.094$&$0.197$  &&$0.926$&$0.089$&$0.193$  \\
    & PatchFusion~\cite{patchfusion} 
    &$0.887$&$0.102$&$0.211$  &&$0.908$&$0.095$&$0.212$  &&$0.881$&$0.116$&$0.258$  \\
    & Ours
    &$\textbf{0.913}$&$\textbf{0.096}$&$\textbf{0.202}$  &&$0.908$&$\textbf{0.091}$&$\textbf{0.197}$  &&$\textbf{0.933}$&$\textbf{0.083}$&$\textbf{0.189}$  \\
    \bottomrule
    \end{tabular}
    }
\end{center}
\vspace{-6pt}
\caption{\textbf{Comparisons with two-stage methods.} PatchFusion~\cite{patchfusion} only adopts ZoeDepth~\cite{zoedepth} as the fixed baseline, while other approaches are pluggable for different depth predictors~\cite{midas,leres,zoedepth}.}
\vspace{-10pt}
\label{tab:tab2stage}
\end{table*}

\subsection{Comparisons with Other Depth Refinement Approaches}
\noindent \textbf{Comparisons with One-stage Methods.} For fair comparisons, we evaluate one-stage~\cite{layerrefine,graph-GDSR,gbdf} and two-stage tile-based~\cite{boostdepth,patchfusion} approaches separately. The one-stage methods predict refined depth based on the whole image. \sx{} conducts one-stage refinement without the coarse-to-fine manner during inference. Comparisons on Middlebury2021~\cite{middle}, Multiscopic~\cite{multiscopic}, and Hypersim~\cite{hypersim} are shown in~\reftab{}~\ref{tab:tab1}. As prior arts~\cite{layerrefine,graph-GDSR,gbdf}, we use three depth predictors MiDaS~\cite{midas}, LeReS~\cite{leres}, and ZoeDepth~\cite{zoedepth}. Regardless of which depth predictor is adopted, \sx{} outperforms the previous one-stage methods~\cite{layerrefine,graph-GDSR,gbdf} in depth accuracy and edge quality on the three datasets~\cite{middle,multiscopic,hypersim}. For instance, our method shows $6.6\%$ and $20.7\%$ improvements over Kim~\textit{et al.}~\cite{layerrefine} for $\text{REL}$ and $\text{ORD}$ with MiDaS~\cite{midas} on Middlebury2021~\cite{middle}, showing the efficacy of our self-distillation paradigm.

\noindent \textbf{Comparisons with Two-stage Tile-based Methods.} Two-stage tile-based methods~\cite{boostdepth,patchfusion} conduct local refinement on numerous patches based on the global refined depth. \sx{} moves away from the tile-based manner and utilizes coarse-to-fine edge refinement to further improve edges and details. As in \reftab{}~\ref{tab:tab2stage}, \sx{} with the coarse-to-fine manner shows obvious advantages. For example, compared with the recent advanced PatchFusion~\cite{patchfusion}, \sx{} achieves $5.2\%$ and $26.7\%$ improvements for $\delta_1$ and $\text{ORD}$ with ZoeDepth~\cite{zoedepth} on Hypersim~\cite{hypersim}. To be mentioned, PatchFusion~\cite{patchfusion} uses ZoeDepth~\cite{zoedepth} as the fixed baseline, whereas \sx{} is readily pluggable for various depth predictors~\cite{midas,leres,zoedepth}.

\begin{figure*}[!t]
\centering
\includegraphics[width=0.95\textwidth,trim=0 0 0 0,clip]{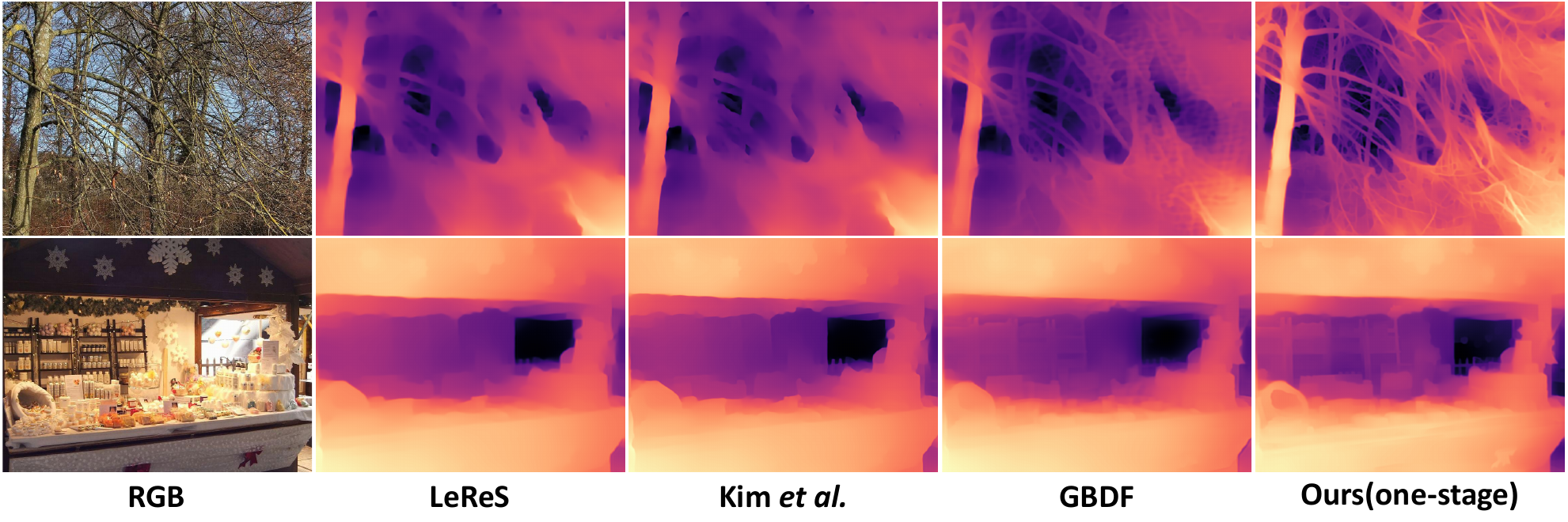}
\vspace{-6pt}
\caption{
\textbf{Qualitative comparisons of one-stage methods on natural scenes.}  LeReS~\cite{leres} is used as the depth predictor. \sx{} predicts sharper depth edges and more meticulous details than prior arts~\cite{gbdf,layerrefine}, \textit{e.g.}, fine-grained predictions of intricate branches. Better viewed when zoomed in.}
\label{fig:1stage}
\vspace{-10pt}
\end{figure*}

\begin{figure*}[!t]
\centering
\includegraphics[width=0.95\textwidth,trim=0 0 0 0,clip]{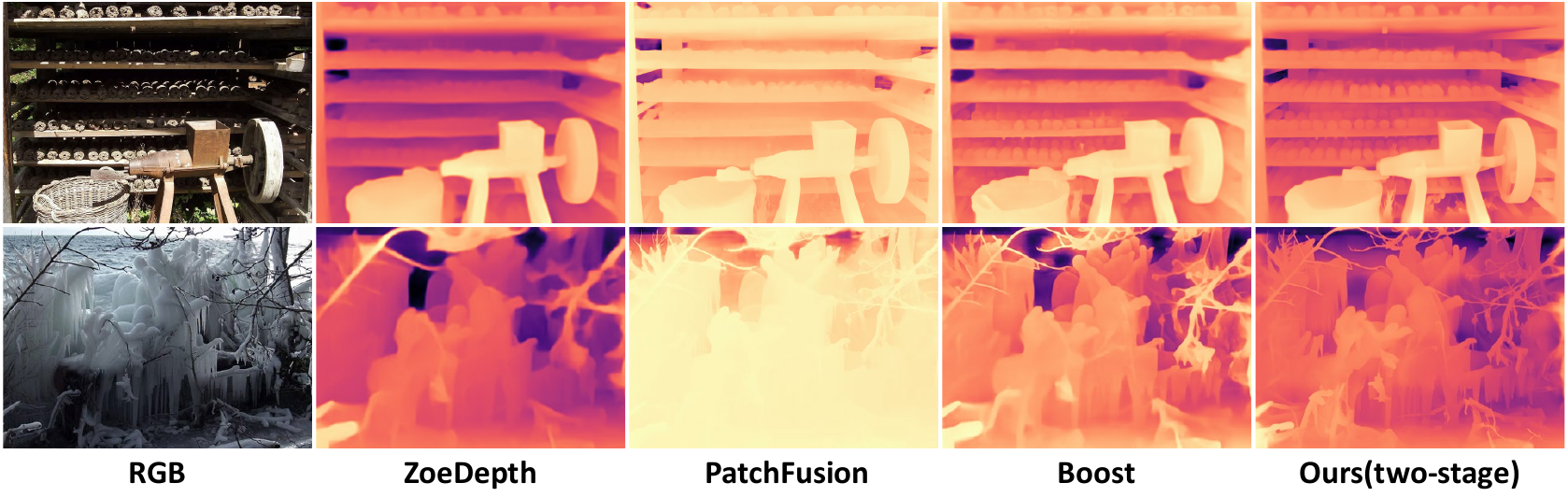}
\vspace{-6pt}
\caption{
\textbf{Qualitative comparisons of two-stage methods on natural scenes.} ZoeDepth~\cite{zoedepth} is adopted as the depth predictor. The \sx{} with coarse-to-fine edge refinement can predict more accurate depth edges and more consistent spatial structures than the tile-based methods~\cite{patchfusion,boostdepth}.}
\label{fig:2stage}
\vspace{-15pt}
\end{figure*}

\begin{figure}[t]
    \centering
    \begin{minipage}{0.6\textwidth}
        \centering
        \resizebox{1.0\textwidth}{!}{
        \begin{tabular}{llcccccccccccc}
        \toprule
        \multirow{2.5}{*}{Method}  &
        \multicolumn{4}{c}{DIML} & &
        \multicolumn{4}{c}{DIODE} \\
        \cmidrule{2-5} \cmidrule{7-10}
        &$\delta_1\!\!\uparrow$ & $\text{REL}\!\downarrow$ & $\text{ORD}\!\downarrow$ & $\text{D}^\text{3}\text{R}\!\downarrow$ & &
        $\delta_1\!\!\uparrow$ & $\text{REL}\!\downarrow$ & $\text{ORD}\!\downarrow$ &$\text{D}^\text{3}\text{R}\!\downarrow$ \\
        \midrule
        LeReS~\cite{leres}    
        &$0.902$&$0.101$&$0.242$&$0.284$  &&$0.892$&$0.105$&$0.324$&$0.685$  \\
        Kim \textit{et al.}~\cite{layerrefine}   
        &$0.902$&$0.100$&$0.243$&$0.301$  &&$0.889$&$0.105$&$0.325$&$0.713$  \\
        Graph-GDSR~\cite{graph-GDSR}   
        &$0.901$&$0.101$&$0.243$&$0.300$  &&$0.890$&$0.104$&$0.326$&$0.690$  \\
        GBDF~\cite{gbdf}    
        &$0.906$&$0.100$&$0.239$&$0.267$  &&$0.894$&$0.105$&$0.322$&$0.673$  \\
        Boost~\cite{boostdepth}    
        &$0.897$&$0.108$&$0.274$&$0.438$  &&$0.892$&$0.105$&$0.343$&$0.640$  \\
        Ours     
        &$\textbf{0.926}$&$\textbf{0.098}$&$\textbf{0.221}$&$\textbf{0.220}$  &&$\textbf{0.900}$&$\textbf{0.098}$&$\textbf{0.293}$&$\textbf{0.637}$  \\
        \bottomrule
        \end{tabular}}
        \captionof{table}{\textbf{Comparisons of model generalizability.} We conduct zero-shot evaluations on DIML~\cite{diml} and DIODE~\cite{diode} datasets with diverse in-the-wild scenarios to compare the generalization capability. We adopt LeReS~\cite{leres} as the depth predictor for all the compared methods in this experiment.}
    \label{tab:generalize}
    \end{minipage}
    \hfill 
    \begin{minipage}{0.37\textwidth}
        \flushright
        \includegraphics[width=1.0\linewidth]{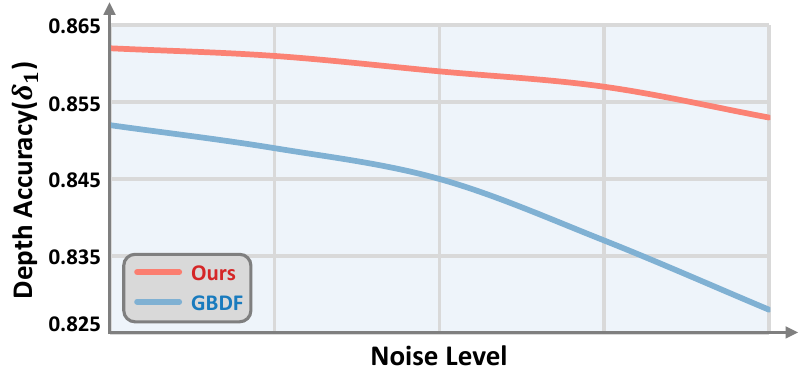} 
        \captionof{figure}{\textbf{Robustness against noises.} X-axis shows noise level of $\epsilon_{\text{cons}}+\epsilon_{\text{edges}}$. With higher noises, our \sx{} is more robust with less performance degradation than the prior GBDF~\cite{gbdf}.}
    \label{fig:noiseexp}
    \end{minipage}
    \vspace{-10pt}
\end{figure}

\noindent\textbf{Generalization Capability on Natural Scenes.}
We prove the superior generalization capability of \sx{}. In this experiment, we adopt LeReS~\cite{leres} as the depth predictor. DIML~\cite{diml} and DIODE~\cite{diode} datasets are used for zero-shot evaluations, considering their diverse in-the-wild indoor and outdoor scenarios. As in~\reftab{}~\ref{tab:generalize}, \sx{} shows at least $5.7\%$ and $9.0\%$ improvements for $\text{REL}$ and $\text{ORD}$ on DIODE~\cite{diode}. On DIML~\cite{diml} dataset, our approach improves $\text{D}^3\text{R}$, $\text{ORD}$, and $\delta_1$ by over $17.6\%$, $7.5\%$, and $2.0\%$. The convincing performance proves our strong robustness and generalizability, indicating the efficacy of our \bosong{} modeling and self-distilled training paradigm.

\noindent \textbf{Qualitative Comparisons.} We present visual comparisons of one-stage methods~\cite{layerrefine,gbdf} on natural scenes in \reffig{}~\ref{fig:1stage}. With our low-noise depth edge representation and edge-based guidance, \sx{} predicts sharper depth edges and details, \textit{e.g.}, the fine-grained predictions of intricate branches.

The visual results of two-stage approaches~\cite{boostdepth,patchfusion} are shown in \reffig{}~\ref{fig:2stage}. Due to the excessive fusion of detailed information, tile-based methods~\cite{boostdepth,patchfusion} produce structure disruption, depth inconsistency, or even noticeable artifacts, \textit{e.g.}, disrupted and fuzzy structures of the snow-covered branches. By contrast, \sx{} can predict more accurate depth edges and more consistent spatial structures.

\noindent \textbf{Robustness against noises.} As in \reffig{}~\ref{fig:noiseexp}, we evaluate \sx{} and GBDF~\cite{gbdf} with different levels of input noises. As the noise level increases, our method presents less degradation. The stronger robustness against the $\epsilon_{\text{cons}}$ and $\epsilon_{\text{edges}}$ noises is the essential reason for all our superior performance.

\noindent \textbf{Model Efficiency.} \sx{} achieves higher efficiency.  Two-stage tile-based methods~\cite{boostdepth,patchfusion} rely on complex fusion of extensive patches with heavy computational overhead. Our coarse-to-fine manner noticeably reduces Flops per patch and patch numbers as in \reffig{}~\ref{fig:fig1}. For one-stage methods~\cite{graph-GDSR,gbdf,layerrefine}, \sx{} adopts a more lightweight $\mathcal{N}_r$ with less parameters and faster inference speed over the previous GBDF~\cite{gbdf} and Kim~\textit{et al.}~\cite{layerrefine}. See Appendix~\ref{sec:efficient} for detailed comparisons of model efficiency. 

\subsection{Ablation Studies}
\noindent \textbf{Coarse-to-fine Edge Refinement.} 
In \reftab{}~\ref{tab:ab_s}, we adopt the coarse-to-fine manner with varied iterations. $S=0$ represents one-stage inference. Coarse-to-fine refinement brings more fine-grained edge representations and refined depth. We set $S=3$ for the \sx{} with two-stage inference.

\noindent \textbf{Edge-based Guidance.} In \reftab{}~\ref{tab:ab_loss}, we evaluate the effectiveness of edge-based guidance. $\mathcal{L}_{grad}$ focuses on consistent refinement of depth edges. $\mathcal{L}_{fusion}$ guides the adaptive feature fusion of low- and high-frequency information. With $\mathcal{L}_{gt}$ as the basic supervision of ground truth, adding $\mathcal{L}_{grad}$  and $\mathcal{L}_{fusion}$ improves $\text{D}^3\text{R}$ by $10.0\%$ and REL by $3.2\%$, showing the efficacy of edge-based guidance.

\noindent \textbf{Effectiveness of SDDR Framework.} As in \reftab{}~\ref{tab:ab_data}, we train \sx{} with the same HRWSI~\cite{kexian2020} as GBDF~\cite{gbdf} for fair comparison. Without the combined training data in Appendix~\ref{sec:datatrainval}, \sx{} still improves $\text{D}^3\text{R}$ and ORD by $13.9\%$ and $2.2\%$ over GBDF~\cite{gbdf}, proving our superiority convincingly.

\begin{table}[t]
    \begin{minipage}{0.65\textwidth}
    \resizebox{0.95\textwidth}{!}{
    \begin{subtable}{0.88\textwidth}
        \centering
        \begin{tabular}{lcccc}
        \toprule
        Method  &
        $\delta_1\!\!\uparrow$ & $\text{REL}\!\downarrow$ & $\text{ORD}\!\downarrow$ & $\text{D}^\text{3}\text{R}\!\downarrow$ \\
        \midrule
        $S=0$               &$0.859$&$0.125$&$0.313$&$0.235$  \\
        $S=1$               &$0.860$&$0.122$&$0.309$&$0.223$  \\
        $S=2$               &$0.860$&$0.120$&$0.307$&$0.219$  \\
        $S=3$               &$\textbf{0.862}$&$\textbf{0.120}$&$\textbf{0.305}$&$\textbf{0.216}$  \\
        \bottomrule
        \end{tabular}
        \captionsetup{font=Large}
        \caption{Coarse-to-fine Edge Refinement} 
        \label{tab:ab_s}
    \end{subtable}%
    \hspace{-0.5cm}
    \begin{subtable}{0.88\textwidth}
        \centering
        \begin{tabular}{lcccccc}
        \toprule
        $\mathcal{L}_{gt}$ & $\mathcal{L}_{grad}$ & $\mathcal{L}_{fusion}$ & $\delta_1\!\!\uparrow$ & $\text{REL}\!\downarrow$ & $\text{ORD}\!\downarrow$ & $\text{D}^\text{3}\text{R}\!\downarrow$ \\
        \midrule
        \cmark&&&               $0.854$&$0.124$&$0.313$&$0.240$  \\
        \cmark&\cmark&&         $0.858$&$0.122$&$0.307$&$0.220$  \\
        \cmark&&\cmark&         $0.859$&$0.120$&$0.311$&$0.229$  \\
        \cmark&\cmark&\cmark&   $\textbf{0.862}$&$\textbf{0.120}$&$\textbf{0.305}$&$\textbf{0.216}$  \\
        \bottomrule
        \end{tabular}
        \captionsetup{font=Large}
        \caption{Edge-based Guidance}
        \label{tab:ab_loss}
    \end{subtable}}
    \end{minipage}
    \begin{minipage}{0.33\textwidth}
    \flushleft
    \begin{minipage}{0.99\textwidth}
    \resizebox{1.0\textwidth}{!}{
    \centering
    \begin{subtable}{2.0\textwidth}
        \begin{tabular}{lcccccc}
        \toprule
        Method  & Training Data &
        $\delta_1\!\!\uparrow$ & $\text{REL}\!\downarrow$ & $\text{ORD}\!\downarrow$ & $\text{D}^\text{3}\text{R}\!\downarrow$ \\
        \midrule
        GBDF~\cite{gbdf} & HRWSI~\cite{kexian2020}  &$0.852$&$0.122$&$0.316$&$0.258$  \\
        Ours &  HRWSI~\cite{kexian2020}             &$\textbf{0.860}$&$\textbf{0.121}$&$\textbf{0.309}$&$\textbf{0.222}$  \\
        \bottomrule
        \end{tabular}
        \captionsetup{font=Large}
        \caption{Effectiveness} 
        \label{tab:ab_data}
    \end{subtable}}
    \end{minipage}
    \begin{minipage}{0.99\textwidth}
    \vspace{-3pt}
    \flushleft
    \resizebox{1.02\textwidth}{!}{
    \begin{subtable}{1.7\textwidth}
        \begin{tabular}{lccccc}
        \toprule
        Method  &
        $\delta_1\!\!\uparrow$ & $\text{REL}\!\downarrow$ & $\text{ORD}\!\downarrow$ & $\text{D}^\text{3}\text{R}\!\downarrow$ \\
        \midrule
        GBDF~\cite{gbdf}    &$0.852$&$0.122$&$0.316$&$0.258$  \\
        GBDF ($w/\,G_S$)     &$\textbf{0.858}$&$\textbf{0.122}$&$\textbf{0.307}$&$\textbf{0.230}$  \\
        \bottomrule
        \end{tabular}
        \captionsetup{font={large}}
        \caption{Transferability}
        \label{tab:ab_label}
    \end{subtable}}%
    \end{minipage}
    \end{minipage}

\caption{\textbf{Ablation Study.} All ablations are on Middlebury2021~\cite{middle} with depth predictor LeReS~\cite{leres}.}
\vspace{-18pt}
\end{table}

\noindent \textbf{Transferability.} We hope our depth edge representation $G_S$ can be applicable to other depth refinement models. Therefore, in \reftab{}~\ref{tab:ab_label}, we directly train GBDF~\cite{gbdf} combining the depth edge representation produced by the trained \sx{}. The depth accuracy and edge quality are improved over the original GBDF~\cite{gbdf}, indicating the transferability of $G_S$ in training robust refinement models. 

\section{Conclusion}
In this paper, we model the depth refinement task as a \bosong{} problem. To enhance the robustness against local inconsistency and \noiseedge{}, we propose \framework{} (\sx{}) framework. With the low-noise depth edge representation and guidance, \sx{} achieves both consistent spatial structures and meticulous depth edges. Experiments showcase our stronger generalizability and higher efficiency over prior arts. The \sx{} provides a new perspective for depth refinement in future works. Limitations and broader impact are discussed in Appendix~\ref{sec:broader}.

\paragraph{Acknowledgement} This work is supported by the National Natural Science Foundation of China under Grant No.\  62406120.

\medskip
{
\small
\bibliographystyle{plain}
\bibliography{main}
}


\clearpage
\appendix

\section{More Details on \sx{} Framework}
\label{sec:A}
\subsection{Depth Edge Representation}
\label{sec:app_representation}

\begin{figure*}[!t]
\centering
\includegraphics[width=1.0\textwidth,trim=0 0 0 0,clip]{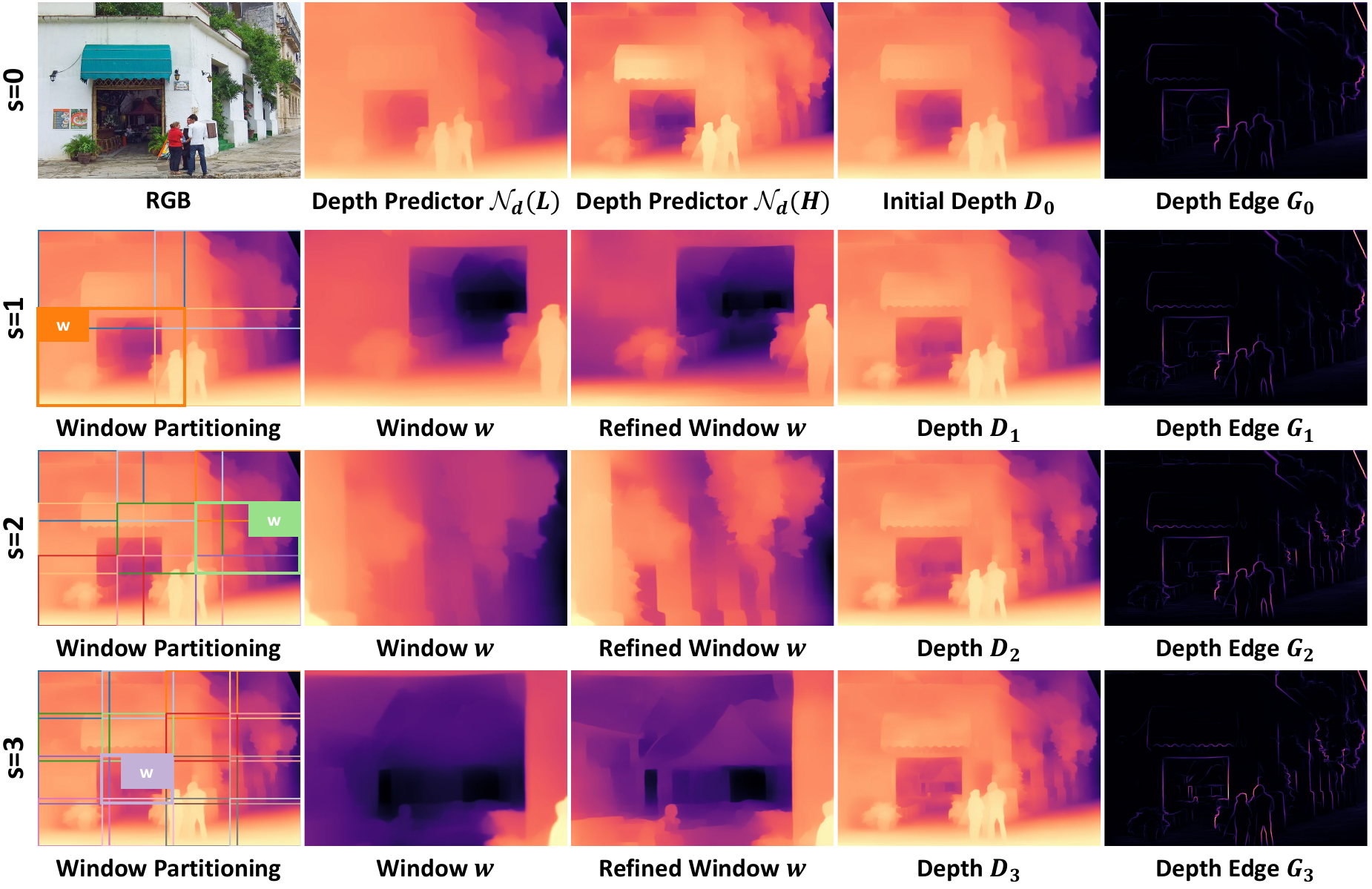}
\vspace{-15pt}
\caption{
\textbf{Visualizations of coarse-to-fine edge refinement.} We present coarse-to-fine results of steps $s=0,1,2,3$. For $s=0$, we showcase the low- and high-resolution predictions $\mathcal{N}_d(L)$ and $\mathcal{N}_d(H)$ of the depth predictor, along with the initial refined depth $D_0$ and edge representation $G_0$. For $s=1,2,3$, we present the window partitioning on the previous $D_{s-1}$, the previous depth $D_{s-1}^w$ on a certain window $w$, refined depth $D_{s}^w$ on the window $w$, refined depth $D_{s}$ of the whole image, and the depth edge representation $G_s$ generated on the current step.}
\label{fig:supp_generate}
\end{figure*}

\noindent \textbf{Coarse-to-fine Edge Refinement.}
In Sec.~3.2, line 169 of main paper, we propose the coarse-to-fine edge refinement to generate accurate and fine-grained depth edge representation $G_S$. Here, we provide visualizations of the refinement process in \reffig{}~\ref{fig:supp_generate}. For the initial global refinement stage $s=0$, we showcase the results of the depth predictor at low and high inference resolutions, \textit{i.e.}, $\mathcal{N}_d(L)$ and $\mathcal{N}_d(H)$. Our refined depth $D_0$ presents both depth consistency and details. For $s=1, 2, 3$, the refined depth maps and edge representations are noticeably improved with finer edges and details. The final depth edge representation $G_S$ ($S=3$) with lower \noisecon{} and \noiseedge{} is utilized as pseudo-label for the self-distillation training process. 

\noindent \textbf{Adaptive Resolution Adjustment.} Adaptive resolution adjustment is applied to the low and high-resolution input $L$ and $H$. We denote the resolutions of $L$ and $H$ as $l$ and $h$, which play a crucial role in refined depth and need to be chosen carefully. Higher resolutions will bring finer details but could lead to inconsistent depth structures due to the limited receptive field of $\mathcal{N}_d$. Previous works~\cite{layerrefine,boostdepth,patchfusion} upscale images or patches to excessively high resolutions for more details, resulting in evident artifacts in their refined depth maps with higher levels of inconsistency noises $\epsilon_{\text{cons}}$. On the other hand, if $h$ is too low, edge and detailed information cannot be sufficiently preserved in $\mathcal{N}_d(H)$, leading to exacerbation of edge deformation noise $\epsilon_\text{edge}$ with blurred details in the refined depth. Such errors and artifacts are unacceptable in depth edge representations for training models. Therefore, we adaptively adjust resolutions $l$ and $h$, considering both the density of depth edges and the training resolution of depth predictor $\mathcal{N}_d$.

For image window $I_s^w$, we generally set the low-resolution input $L_s^w$ as the training resolution $\hat{r}$ of $\mathcal{N}_d$. If we denote the original resolution of $I_s^w$ as $r_s^w$, \sx{} adaptively adjusts the high resolution $h_s^w$ for the certain window as follows:
\begin{equation}
    h_s^w = \textit{mean}(\hat{r},r_s^w) * 
     \frac{\textit{mean}(|\nabla{\mathcal{N}_d(L_s^w)}|)}{\alpha} * \frac{\textit{mean}(|\nabla{D_{s-1}^{w}}|)}{\textit{mean}(|\nabla{D_{s-1}}|)}\,, \\
     \label{eq:res}
\end{equation}
where $\alpha$ is a priori parameter for depth predictor $\mathcal{N}_d$, averaging the gradient magnitude of the depth annotations on its sampled training data. The second term embodies adjustments according to depth edges. Assuming $\textit{mean}(|\nabla\mathcal{N}_d(L_s^w)|)<\alpha$, it indicates that the current window area contains lower edge intensity or density than the training data of $\mathcal{N}_d$. In this case, we will appropriately decrease $h_s^w$ from $\textit{mean}(\hat{r},r_s^w)$ to maintain the similar density of detailed information as the training stage of the depth predictor. The third term portrays adjustments based on the discrepancy of edge intensity between the window area and the whole image. To be mentioned, for the generation of the initial edge representation $G_0$, the third term is set to ineffective as one. $L_0^w$ is equivalent to $L$ with the whole image as the initial window $w$.

\begin{figure*}[!t]
\centering
\includegraphics[width=1.0\textwidth,trim=0 0 0 0,clip]{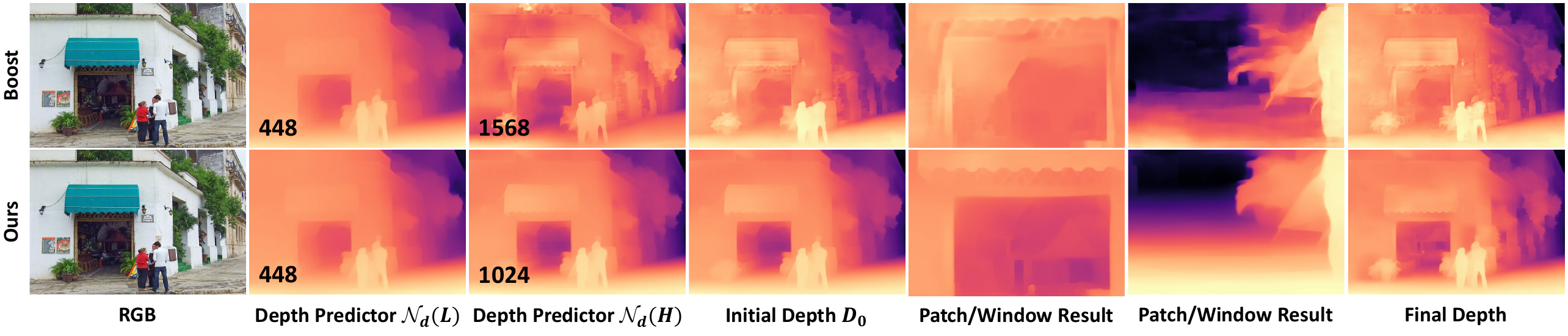}
\vspace{-15pt}
\caption{
\textbf{Adaptive resolution adjustment.} We compare the effects of inference resolutions with Boost~\cite{boostdepth}. The numbers in the corner of the second and third columns represent the chosen inference resolution. We relieve the artifacts in Boost~\cite{boostdepth} by adaptive resolution adjustment.}
\label{fig:supp_res}
\end{figure*}

We present visual results with different resolutions to prove the effectiveness of our design. As shown in \reffig{}~\ref{fig:supp_res}, considering the training data distribution and the edge density, the inference resolution is adaptively adjusted to a smaller one compared to Boost~\cite{boostdepth} ($1024$ versus $1568$). In this way, our \sx{} achieves better depth consistency and alleviates the artifacts produced by prior arts~\cite{gbdf,boostdepth}.

\subsection{Edge-based Guidance}
\begin{figure*}[!t]
\centering
\includegraphics[width=1.0\textwidth,trim=0 0 0 0,clip]{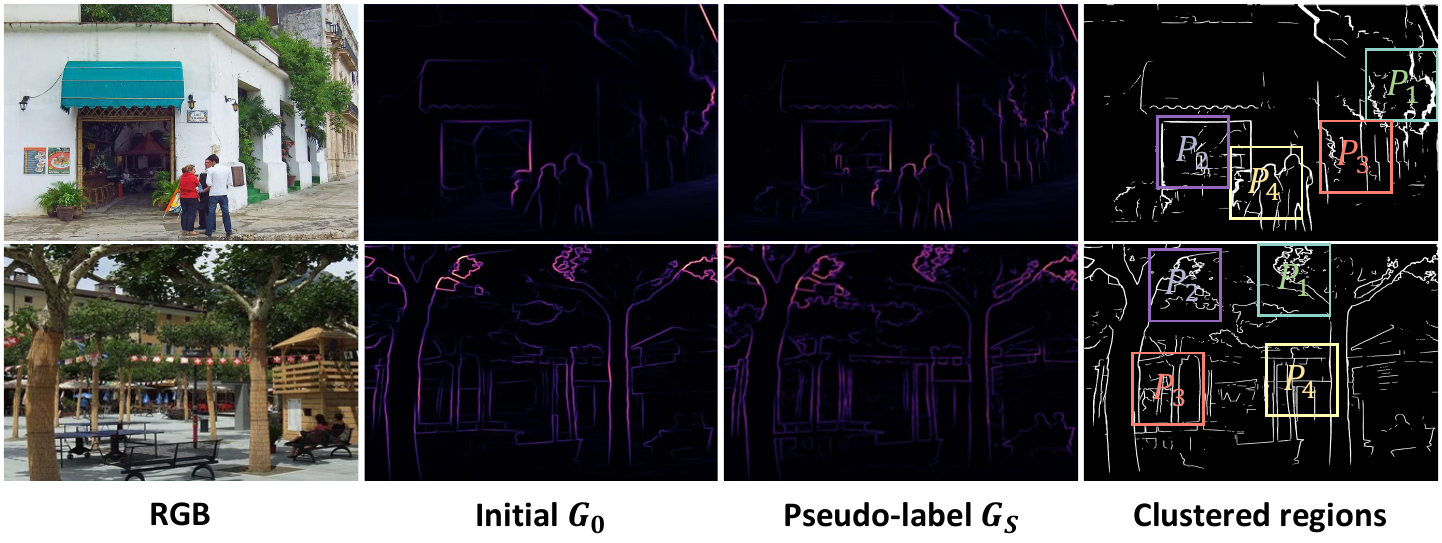}
\vspace{-15pt}
\caption{
\textbf{Edge-guided gradient error.} $\mathcal{L}_{grad}$ focuses on high-frequency areas $P_n$ extracted by clustering with more details. The flat regions are not constrained to preserve depth consistency.}
\vspace{-10pt}
\label{fig:supp_gradloss}
\end{figure*}

\noindent \textbf{Edge-guided Gradient Error.}
In line 192, Sec~3.3 of the main paper, we mention that we use clustering to obtain several high-frequency local regions to compute our \lossgrad{}. Here, we elaborate on the details. K-means clustering~\cite{kmeans} is utilized to obtain the edge-dense areas. Specifically, we binarize the edge pseudo-label, setting the top $5\%$ pixels to one and the rest to zero. Next, we employ k-means clustering on the binarized labels to get several edge-dense areas with the centroid value as one. The clustered areas are shown in the fourth column of the \reffig{}~\ref{fig:supp_gradloss}. Our \lossgrad{} supervises these high-frequency regions to improve depth details. The depth consistency in flat areas can be preserved without the constraints of depth edges.

\begin{figure*}[!t]
\centering
\includegraphics[width=1.0\textwidth,trim=0 0 0 0,clip]{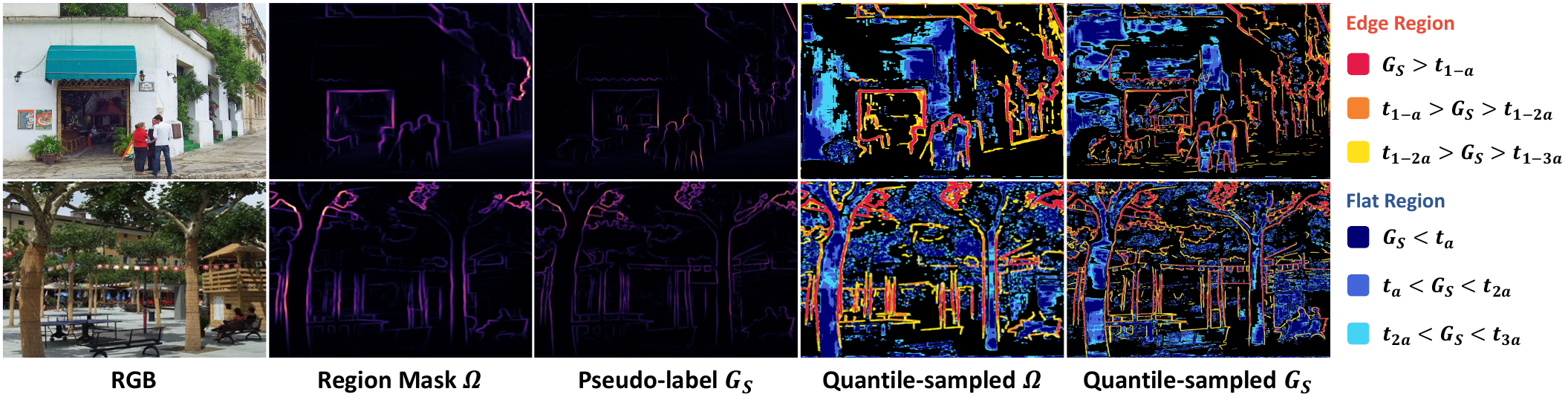}
\vspace{-15pt}
\caption{
\textbf{Edge-based fusion error.}  We present the region mask $\Omega$ and pseudo-label $G_S$ before and after quantile sampling. Different colors on the right represent the range of pixel values. Guiding the $\Omega$ with $G_S$ ensures that our model can predict balanced consistency and details by the simple one-stage inference. The use of $\Omega$ as a learnable soft mask achieves more fine-grained integration on the feature level, enhancing the accuracy of $\mathcal{N}_r$. This also leads to more accurate edge representation $G_S$ in the iterative coarse-to-fine refinement process.}
\label{fig:supp_lossweight}
\end{figure*}

\noindent \textbf{Edge-based Fusion Error.}
The proposed \lossweight{} aligns the data distribution of the learnable region mask $\Omega$ and the pseudo-label $G_S$ by quantile sampling (Sec~3.3, line 205, main paper). Here, we provide additional visualizations for intuitive understanding. As shown in \reffig{}~\ref{fig:supp_lossweight}, we visualize the soft region mask $\Omega$ of high-frequency areas and the pseudo-label $G_S$ with the same color map in the second and third columns. The regions highlighted in $G_S$ with stronger depth edges and more detailed information naturally correspond to larger values in $\Omega$ to emphasize features from high-resolution inputs. We perform quantile sampling on $\Omega$ and $G_S$, as depicted in the fourth and fifth columns. The legends on the right indicate the percentile ranking of the pixel values in the whole image. Our \lossweight{} supervises that $\Omega$ and $G_S$ have consistent distribution for each color. In this way, $\Omega$ tends to have smaller values in flat regions for more information from low-resolution input, while the opposite is true in high-frequency regions. This is advantageous for the model to balance the depth details and spatial structures.

\begin{figure*}[t]
\centering
\includegraphics[width=1.0\textwidth,trim=0 0 0 0,clip]{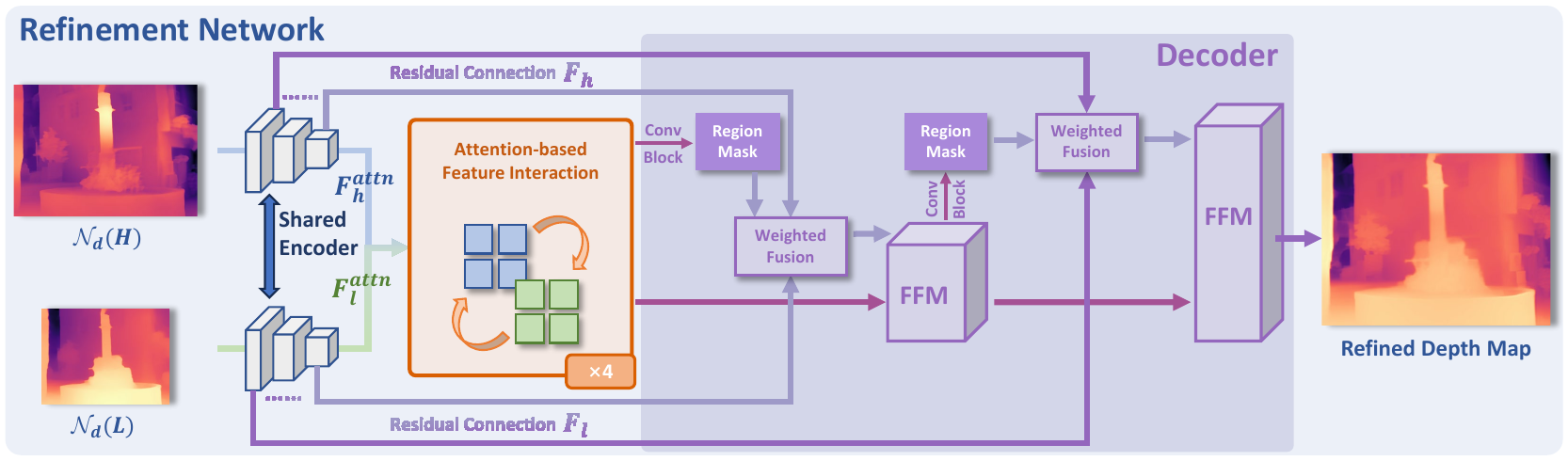}
\vspace{-15pt}
\caption{\textbf{Architecture of refinement network.} Some decoder layers are omitted for simplicity.}
\label{fig:supp_network}
\vspace{-10pt}
\end{figure*}

\subsection{Refinement Network}
We provide the detailed model architecture of the \network{} $\mathcal{N}_r$. As shown in \reffig{}~\ref{fig:supp_network}, the \network{} adopts the U-Net architecture similar to prior arts~\cite{boostdepth,gbdf,layerrefine}. The depth maps from the depth predictor $\mathcal{N}_d$ predicted in different resolutions are up-sampled to a unified input size. A shared Mit-b0~\cite{segformer} serves as the encoder to extract feature maps of different resolutions. The decoder gradually outputs the refined depth map with feature fusion modules (FFM)~\cite{FFM1,FFM2} and skip connections. We make two technical improvements to the refinement network, including attention-based feature interaction and adaptive weight allocation.

\noindent \textbf{Attention-based Feature Interaction.}
To predict refined depth maps in high resolution (\textit{e.g.}, $2048\times2048$), prior arts~\cite{boostdepth,gbdf,layerrefine} adopt a U-Net with numerous layers (\textit{e.g.}, $10$ layers or more) as the \network{} for sufficient receptive field. This leads to heavy computational overhead. In our case, we leverage the self-attention mechanism~\cite{transformer} to address this issue. 

The features of low- and high-resolution inputs extracted by the encoder~\cite{segformer} are denoted as $F^{attn}_l$ and $F^{attn}_h$. We stack $F^{attn}_l$ and $F^{attn}_h$ to obtain $F^{in}$ for attention calculation. Positional embeddings~\cite{vaswani2017attention} $PE_x$, $PE_y$ are added to $F^{in}$ for the height and width dimensions. An additional $PE_f$ is used to distinguish the low- and high-resolution inputs. The attention-based feature interaction process can be expressed as follows: 
\begin{equation}
\begin{gathered}
F^{in}=\operatorname{Stack}(F^{attn}_l, F^{attn}_h)+PE_x+PE_y+PE_f\,, \\
K=W^k \cdot F^{in}, Q=W^q \cdot F^{in}, V=W^v \cdot F^{in}, \\
F^{out}=\operatorname{Softmax}\!\left(K^T Q / {\sqrt{d}}\right)\!V + F^{in}\,. \\
\end{gathered}
\end{equation}
Four attention layers are included in $\mathcal{N}_r$. The interacted feature $F^{out}$ is fed to the decoder to predict refined depth. Attention-based feature interaction achieves large receptive field with fewer layers, reducing model parameters and improving efficiency.

\noindent \textbf{Adaptive Weight Allocation.} The \network{} adopts adaptive weight allocation for the fusion of low- and high-resolution features with the learnable mask $\Omega$. In each decoder layer, the feature go through a convolutional block to generate $\Omega$ with a single channel. The fused features $F$ (line 212, main paper) and the feature from the previous layer are fused by the FFM module~\cite{FFM2}.

\subsection{Noise Implementation.}
\label{sec:NoiseImplementation.}
For our \noisecon{}, we segment the ideal depth $D^*$ into regular patches of size $64\times64$, with an overlap of half the patch size. Considering the depth discontinuities on the edges, instead of applying a linear transformation to the entire patch, we extract the edges from $D^*$ and apply a linear transformation to each connected domain to simulate the local depth inconsistency. For edge deformation noise, we first down-sample $D^*$ to the inference resolution and then restore it to the original resolution. Subsequently, we optimize a certain number of Gaussian distributions around the edges of $D^*$ to fit the edge deformation and blurring.

The local inconsistency noise and edge deformation noise can effectively model the degradation of network prediction results compared to ideal depth maps. An additional experiment on the Middlebury2021~\cite{middle} dataset also proves this point. We optimize the \noisecon{} with the least squares method and $50@000$ position-constrained Gaussian distributions as edge deformation noise by gradient descent. The PSNR between the noisy depth $(D^*+\epsilon_\text{cons}+\epsilon_\text{edge})$ and model predicted depth $D$ is over $40$ dB, which indicates that the difference between $D$ and $(D^*+\epsilon_\text{cons}+\epsilon_\text{edge})$ is very small. The result further demonstrates that the noises can accurately model depth prediction errors (\refequ{}~\ref{eq:11111}, main paper), similar to the visualizations in \reffig{}~\ref{fig:noise} of the main text.

\subsection{Broader Impacts and Limitations} 
\label{sec:broader}
Although \sx{} works well in general, it still has limitations. For example, more advanced mechanisms and structures can be explored for the refinement network in future work. For inputs under conditions with specular surfaces, low light, or weak textures, the depth predictor tends to yield sub-optimal results. Although \sx{} improves upon these results, the outcomes are still not perfect. Our approach exclusively utilizes publicly available datasets during the training process, thereby having no broad societal impact, not involving AI ethics, and not involving any privacy-sensitive data.

\section{Detailed Experimental Settings}
\label{sec:B}
\subsection{Datasets}
\label{sec:datatrainval}
\noindent \textbf{Evaluation Datasets.} We use five different benchmarks with diverse scenarios for comparisons. The descriptions of our evaluation datasets are as follows:
\begin{itemize}[leftmargin=*]
\item[$\bullet$] \textbf{Middlebury2021}~\cite{middle} comprises 48 RGB-D pairs from 24 real indoor scenes for evaluating stereo matching and depth refinement models. Each image in the dataset is annotated with dense $1920 \times 1080$ disparity maps. We use the whole set of Middlebury2021~\cite{middle} for testing.

\item[$\bullet$] \textbf{Multiscopic}~\cite{multiscopic} includes a test set with $100$ synthetically generated indoor scenes. Each scene consists of RGB images captured from $5$ different viewpoints, along with corresponding disparity annotations. The resolution of images is $1280 \times 1080$. We adopt its official test set for testing.

\item[$\bullet$] \textbf{Hypersim}~\cite{hypersim} is a large-scale synthetic dataset. In our experiment, we follow the test set defined by GBDF~\cite{gbdf} for fair comparison, utilizing tone-mapped $286$ images generated by their released code. Evaluation is performed using the corresponding $1024 \times 768$ depth annotations.

\item[$\bullet$] \textbf{DIML}~\cite{diml} contains RGB-D frames from both Kinect v2~\cite{kinect} and Zed stereo camera with different resolutions. We conduct the generalization evaluation using the official test set, which includes real indoor and outdoor scene images along with corresponding high-resolution depth annotations.

\item[$\bullet$] \textbf{DIODE}~\cite{diode} contains high-quality $1024 \times 768$ LiDAR-generated depth maps of both indoor and outdoor scenes. We use the whole validation set (771 images) for generalization testing.
\end{itemize}

\noindent \textbf{Training Datasets.} 
Our training data is sampled from diverse datasets, which can be categorized into synthetic and natural-scene datasets. The synthetic datasets consist of TartanAir~\cite{tata}, Irs~\cite{irs}, UnrealStereo4K~\cite{UnrealStereo4K} and MVS-Synth~\cite{MVS-Synth}. Among these, the resolutions of TartanAir~\cite{tata} and Irs~\cite{irs} are below 1080p, while MVS-Synth~\cite{MVS-Synth} and UnrealStereo4K~\cite{UnrealStereo4K} reach resolutions of 1080p and 4k, respectively. Irs~\cite{irs} and MVS-Synth~\cite{MVS-Synth} contain limited types of scenes, whereas others include both indoor and outdoor scenes, some of which~\cite{tata,UnrealStereo4K} present challenging conditions like poor lighting. To enhance the generalization to natural scenes, we also sample from four high-resolution real-world datasets, Holopix50K~\cite{holopix50k}, iBims-1~\cite{ibims}, WSVD~\cite{wsvd}, and VDW~\cite{nvds}. IBims-1~\cite{ibims} contains a small number of indoor scenes but provides high-precision depth annotations from the capturing device. The remaining three datasets include large-scale diverse scenes, but their depth annotations, obtained from stereo images~\cite{flownet2}, lack ideal edge precision.

\subsection{Training Recipe}
\label{sec:TrainingRecipe}
We leverage diverse training data to achieve strong generalizability. For each epoch, we randomly choose $20@000$ images from natural-scene data~\cite{holopix50k,nvds,wsvd,ibims} and $20@000$ images from synthetic datasets~\cite{tata,irs,UnrealStereo4K,MVS-Synth}. For each sample, we adopt similar data processing and augmentation as GBDF~\cite{gbdf}. To enhance training stability, we first train $\mathcal{N}_r$ for one epoch only with $\mathcal{L}_{gt}$. In the next two epochs, we involve $\mathcal{L}_{grad}$ and $\mathcal{L}_{fusion}$ for self-distillation. The $a$ and $N_w$ in $\mathcal{L}_{fusion}$ are set to $0.02$ and $4$. The learning rate is $1e-4$. $\lambda_1$ and $\lambda_2$ in \refequ{}~\ref{eq:lossall} are $0.5$ and $0.1$. All training and inference are conducted on a single NVIDIA A6000 GPU.

\subsection{Evaluation Metrics}
\noindent \textbf{Depth Accuracy.}
$M$ denotes numbers of pixels with valid depth annotations, while $d_i$ and $d_i^*$ are estimated and ground truth depth of pixel $i$. We adopt the widely-used depth metrics as follows:
\begin{itemize}[leftmargin=*]
\item[$\bullet$] \textbf{Absolute relative error (Abs Rel):} $\frac{1}{|M|} \sum_{d \in M}\left|d-d^*\right| / d^* ;$
\item[$\bullet$] \textbf{Square relative error (Sq Rel):} $\frac{1}{|M|} \sum_{d \in M}\left\|d-d^*\right\|^2 / d^*$
\item[$\bullet$] \textbf{Root mean square error (RMSE):} $\sqrt{\frac{1}{|M|} \sum_{d \in M}\left\|d-d^*\right\|^2} ;$
\item[$\bullet$] \textbf{Mean absolute logarithmic error (log$\mathbf{_{10}}$):} $\frac{1}{|M|} \sum_{d \in M}\left|\log\left(d\right)-\log\left(d^*\right)\right| ;$
\item[$\bullet$] \textbf{Accuracy with threshold t:} Percentage of $d_i$ such that $\\max(\frac{d_i}{d_i^*},\frac{d_i^*}{d_i}) = \delta<t\in\left[1.25, 1.25^2, 1.25^3\right]\,.$ 
\end{itemize}

\noindent \textbf{Edge Quality.}
For the edge quality, we follow prior arts~\cite{boostdepth,gbdf,kexian2020} to employ the ordinal error ($\text{ORD}$) and depth discontinuity disagreement ratio ($\text{D}^\text{3}\text{R}$). The ORD metric is defined as:
\begin{equation}
\begin{gathered}
ORD=\frac{1}{N}\sum_i\phi(p_{i,0}-p_{i,1}) \,,\\
\phi(p_{i,0}-p_{i,1})=\left\{
\renewcommand{\arraystretch}{1.2}
\begin{array}{ll}
\log\left(1+\exp\left(-l\left(p_{i,0}-p_{i,1}\right)\right)\right),    & l \neq 0 \,,\\
(p_{i,0}-p_{i,1})^2,                                        & l = 0 \,,\\
\end{array}
\right. \\
l=\left\{
\begin{array}{lll}
+1,    && p_{i,0}^*/p_{i,1}^* \geq 1+\tau \,,\\
-1,    && p_{i,0}^*/p_{i,1}^* \leq \frac{1}{1+\tau} \,,\\
0,     && otherwise    \,,\\
\end{array}
\right.
\end{gathered}
\end{equation}
where $p_{i,0}$ and $p_{i,1}$ represent pairs of edge-guided sampling points. $p_{i,0}^*$ and $p_{i,1}^*$ are the ground truth values at corresponding positions. $l$ is used to represent the relative ordinal relationship between pairs of points. $\text{ORD}$ characterizes the quality of depth edges by sampling pairs of points near extracted edges using a ranking loss~\cite{kexian2020}. On the other hand, $\text{D}^\text{3}\text{R}$~\cite{boostdepth} uses the centers of super-pixels computed with the ground truth depth and compares neighboring super-pixel centroids across depth discontinuities. It directly focuses on the accuracy of depth boundaries.

\section{More Experimental Results}
\label{sec:C}
\subsection{Model Efficiency Comparisons.} 
\label{sec:efficient}
In line 277 of the main paper, we mention that our method achieves higher model efficiency than prior arts~\cite{graph-GDSR,gbdf,layerrefine,boostdepth,patchfusion}. Here, we provide detailed comparisons of model efficiency in \reftab{}~\ref{tab:efficient}. For one-stage methods~\cite{graph-GDSR,gbdf,layerrefine}, \sx{} adopts a more lightweight \network{}, reducing model parameters by $12.5$ times than GBDF~\cite{gbdf} and improving inference speeds by $3.6$ times than Kim~\textit{et al.}~\cite{layerrefine}. Compared with two-stage tile-based methods~\cite{boostdepth,patchfusion}, our coarse-to-fine edge refinement reduces the Flops per patch by $50.6$ times and the patch numbers by $5.9$ times than PatchFusion~\cite{patchfusion}.
\begin{table*}[t]
    \addtolength{\tabcolsep}{+1.3pt}
    \begin{center}
    \resizebox{0.57\textwidth}{!}{
    \begin{tabular}{lccc}
    \toprule
    Method & FLOPs ($G$) & Params ($M$) & Time ($s$) \\
    \midrule
    GBDF~\cite{gbdf}   
    & $10.377$ & $201.338$ & $0.112$ \\
    Kim~\textit{et al.}~\cite{layerrefine}   
    & $1138.342$ & $61.371$ & $0.128$ \\
    Graph-GDSR~\cite{graph-GDSR}   
    & $397.355$ & $32.533$ & $0.832$ \\
    Ours (one-stage) & $16.733$ & $16.763$ & $0.035$ \\
    \midrule
    Boost~\cite{boostdepth}
    & $286.13\times63$ & $79.565$ & $2.183$ \\
    PatchFusion~\cite{patchfusion}
    & $810.813\times177$ & $42.511$ & $5.345$ \\
    Ours (two-stage)
    & $16.733\times30$ & $16.763$ & $1.050$ \\
    \bottomrule
    \end{tabular}
    }
\end{center}
\vspace{-8pt}
\caption{
\textbf{Model efficiency.} We evaluate FLOPs, model parameters, and inference time of different methods. The first four rows contain one-stage methods~\cite{gbdf,layerrefine,graph-GDSR}, while the last three rows are for two-stage approaches~\cite{boostdepth,patchfusion}. FLOPs and inference time are tested on a $1024\times 1024$ image with one NVIDIA RTX A6000 GPU. For the two-stage methods~\cite{boostdepth,patchfusion}, their FLOPs are reported by multiplying FLOPs per patch with the required patch numbers for processing the image.}
\label{tab:efficient}
\end{table*}

\begin{figure*}[!t]
\centering
\includegraphics[width=0.55\textwidth,trim=0 0 0 0,clip]{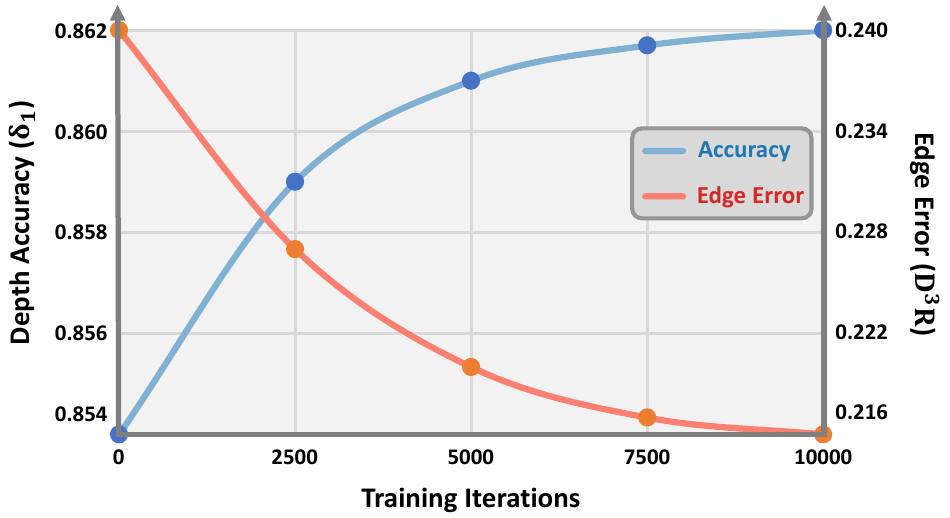}
\vspace{-6pt}
\caption{
\textbf{Iterations for self-distillation.} We report the depth accuracy and edge error metrics of our \sx{} model in the self-distillation training process.}
\label{fig:supp_iter}
\end{figure*}

\begin{table*}[!t]
\centering
\resizebox{0.5\textwidth}{!}{
\begin{tabular}{lccccc}
\toprule
Method & $\delta_1\!\!\uparrow$ & $\text{REL}\!\downarrow$ & $\text{ORD}\!\downarrow$ & $\text{D}^\text{3}\text{R}\!\downarrow$ \\
\midrule
Ours ($w/\,D_S$)   
& $0.855$&$0.129$&$0.317$&$0.237$ \\
Ours ($w/\,G_S$)   
& $\textbf{0.862}$&$\textbf{0.120}$&$\textbf{0.305}$&$\textbf{0.216}$ \\
\bottomrule
\end{tabular}}
\vspace{-4pt}
\caption{\textbf{Formats of Pseudo-labels.} We compare the self-distilled training with refined depth $D_S$ and depth edge representation $G_S$ as pseudo-labels. The experiment is conducted on Middlebury2021~\cite{middle} dataset with LeReS~\cite{leres} as the depth predictor.} 
\vspace{-10pt}
\label{tab:ab_format}
\end{table*}

\begin{table*}[t]
    \begin{center}
    \resizebox{0.85\textwidth}{!}{
    \begin{tabular}{llcccccccccc}
    \toprule
    \multirow{2}{*}{Predictor} &
    \multirow{2}{*}{Method} &
    \multicolumn{7}{c}{Depth} &&
        \multicolumn{2}{c}{Edge} \\
    \cmidrule{3-9} \cmidrule{11-12}
    && Abs Rel$\downarrow$ & Sq Rel$\downarrow$ & RMSE$\downarrow$ & $\log_{10}\!\downarrow$ & $\delta_1\!\uparrow$ & $\delta_2\!\uparrow$ & $\delta_3\!\uparrow$ & & ORD$\downarrow$ & $\text{D}^\text{3}\text{R}\!\downarrow$ \\
    \midrule
    \multirow{5}{*}{MiDaS} & MiDaS~\cite{midas}   
    & $0.117$ & $0.576$ & $3.752$ & $0.052$ & $0.868$ & $0.973$ & $0.992$ &&  
    $0.384$ & $0.334$\\
    & Kim \textit{et al.}~\cite{layerrefine}  
    & $0.120$ & $0.562$ & $\textbf{3.558}$ & $0.053$ & $0.864$ & $0.973$ & $0.994$ && 
    $0.377$ & $0.382$\\
    & Graph-GDSR~\cite{graph-GDSR}  
    & $0.121$ & $0.566$ & $3.593$ & $0.053$ & $0.865$ & $0.973$ & $0.994$ && 
    $0.380$ & $0.398$\\
    & GBDF~\cite{gbdf}   
    & $0.115$ & $0.561$ & $3.685$ & $0.052$ & $0.871$ & $0.973$ & $0.993$ && 
    $0.305$ & $0.237$\\
    & Ours              
    & $\textbf{0.112}$ & $\textbf{0.545}$ & $3.668$ & $\textbf{0.050}$ & $\textbf{0.879}$ & $\textbf{0.979}$ & $\textbf{0.994}$ && 
    $\textbf{0.299}$ & $\textbf{0.220}$\\
    \arrayrulecolor{black}\midrule
    
    \multirow{5}{*}{LeReS} & LeReS~\cite{leres}    
    & $0.123$ & $0.464$ & $3.040$ & $0.052$ & $0.847$ & $0.969$ & $0.992$ &&  
    $0.326$ & $0.359$\\
    & Kim \textit{et al.}~\cite{layerrefine}  
    & $0.124$ & $0.474$ & $3.063$ & $0.052$ & $0.846$ & $0.969$ & $0.992$ && 
    $0.328$ & $0.387$\\
    & Graph-GDSR~\cite{graph-GDSR}  
    & $0.124$ & $0.467$ & $3.052$ & $0.052$ & $0.847$ & $0.969$ & $0.992$ && 
    $0.327$ & $0.373$\\
    & GBDF~\cite{gbdf}   
    & $0.122$ & $\textbf{0.444}$ & $\textbf{2.963}$ & $0.051$ & $0.852$ & $0.969$ & $0.992$ && 
    $0.316$ & $0.258$\\
    & Ours
    & $\textbf{0.120}$ & $0.452$ & $2.985$ & $\textbf{0.050}$ & $\textbf{0.862}$ & $\textbf{0.971}$ & $\textbf{0.993}$ && 
    $\textbf{0.305}$ & $\textbf{0.216}$\\
    \arrayrulecolor{black}\midrule
    
    \multirow{5}{*}{Zoedepth} & Zoedepth~\cite{zoedepth}  
    & $0.104$ & $0.433$ & $2.724$ & $0.043$ & $0.900$ & $0.970$ & $0.993$ && 
    $0.225$ & $0.208$\\
    & Kim \textit{et al.}~\cite{layerrefine}  
    & $0.107$ & $0.469$ & $2.766$ & $0.044$ & $0.896$ & $0.970$ & $0.992$ && 
    $0.228$ & $0.243$\\
    & Graph-GDSR~\cite{graph-GDSR}  
    & $0.103$ & $0.431$ & $2.725$ & $0.044$ & $0.901$ & $0.971$ & $0.993$ && 
    $0.226$ & $0.233$\\
    & GBDF~\cite{gbdf} 
    & $0.105$ & $0.430$ & $2.732$ & $0.044$ & $0.899$ & $0.970$ & $0.993$ && 
    $0.226$ & $0.200$\\
    & Ours
    & $\textbf{0.100}$ & $\textbf{0.406}$ & $\textbf{2.674}$ & $\textbf{0.042}$ & $\textbf{0.905}$ & $\textbf{0.973}$ & $\textbf{0.994}$ && 
    $\textbf{0.218}$ & $\textbf{0.187}$\\  
    \arrayrulecolor{black}\bottomrule
    \end{tabular}
    }
\end{center}
\vspace{-8pt}
\caption{\textbf{Comparisons with one-stage refinement approaches on Middlebury2021.}}
\label{tab:mid21_1s}
\end{table*}

\begin{table*}[t]
    \begin{center}
    \resizebox{0.85\textwidth}{!}{
    \begin{tabular}{llcccccccccc}
    \toprule
    \multirow{2}{*}{Predictor} &
    \multirow{2}{*}{Method} &
    \multicolumn{7}{c}{Depth} &&
        \multicolumn{2}{c}{Edge} \\
    \cmidrule{3-9} \cmidrule{11-12}
    && Abs Rel$\downarrow$ & Sq Rel$\downarrow$ & RMSE$\downarrow$ & $\log_{10}\!\downarrow$ & $\delta_1\!\uparrow$ & $\delta_2\!\uparrow$ & $\delta_3\!\uparrow$ & & ORD$\downarrow$ & $\text{D}^\text{3}\text{R}\!\downarrow$ \\
    \midrule
    \multirow{3}{*}{MiDaS} & MiDaS~\cite{midas}   
    & $0.117$ & $0.576$ & $3.752$ & $0.052$ & $0.868$ & $0.973$ & $0.992$ &&  
    $0.384$ & $0.334$\\
    & Boost~\cite{boostdepth}  
    & $0.118$ & $\textbf{0.544}$ & $3.758$ & $0.053$ & $0.870$ & $\textbf{0.979}$ & $\textbf{0.997}$ &&  
    $0.351$ & $0.257$\\
    & Ours
    & $\textbf{0.115}$ & $0.563$ & $\textbf{3.710}$ & $\textbf{0.052}$ & $\textbf{0.871}$ & $0.973$ & $0.993$ &&  
    $\textbf{0.303}$ & $\textbf{0.248}$\\
    \arrayrulecolor{black}\midrule
    
    \multirow{3}{*}{LeReS} & LeReS~\cite{leres}    
    & $0.123$ & $0.464$ & $3.040$ & $0.052$ & $0.847$ & $0.969$ & $\textbf{0.992}$ &&  
    $0.326$ & $0.359$\\
    & Boost~\cite{boostdepth}
    & $0.131$ & $0.487$ & $3.014$ & $0.054$ & $0.844$ & $0.960$ & $0.989$ &&  
    $0.325$ & $\textbf{0.202}$\\
    & Ours
    & $\textbf{0.123}$ & $\textbf{0.459}$ & $\textbf{3.005}$ & $\textbf{0.052}$ & $\textbf{0.861}$ & $\textbf{0.969}$ & $0.991$ &&  
    $\textbf{0.309}$ & $0.214$\\
    \arrayrulecolor{black}\midrule
    
    \multirow{4}{*}{Zoedepth} & Zoedepth~\cite{zoedepth}  
    & $0.104$ & $0.433$ & $2.724$ & $0.043$ & $0.900$ & $0.970$ & $0.993$ && 
    $0.225$ & $0.208$\\
    & Patchfusion~\cite{patchfusion}  
    & $0.102$ & $0.385$ & $\textbf{2.406}$ & $0.042$ & $0.887$ & $0.977$ & $\textbf{0.997}$ &&  
    $0.211$ & $0.139$\\
    & Boost~\cite{boostdepth}  
    & $0.099$ & $\textbf{0.349}$ & $2.502$ & $0.042$ & $0.911$ & $\textbf{0.979}$ & $0.995$ &&  
    $0.210$ & $0.140$\\
    & Ours
    & $\textbf{0.096}$ & $0.350$ & $2.432$ & $\textbf{0.041}$ & $\textbf{0.913}$ & $0.977$ & $0.995$ &&  
    $\textbf{0.202}$ & $\textbf{0.125}$\\
    \arrayrulecolor{black}\bottomrule
    \end{tabular}
    }
\end{center}
\vspace{-8pt}
\caption{\textbf{Comparisons with two-stage tile-based methods on Middlebury2021. }PatchFusion~\cite{patchfusion} can only adopt
ZoeDepth~\cite{zoedepth} as the fixed baseline, while other approaches are reconfigurable and pluggable for
different depth predictors~\cite{zoedepth,leres,midas}.}
\label{tab:mid21_2s}
\end{table*}

\begin{table*}[!t]
    \begin{center}
    \resizebox{0.85\textwidth}{!}{
    \begin{tabular}{llcccccccccc}
    \toprule
    \multirow{2}{*}{Dataset} &
    \multirow{2}{*}{Method} &
    \multicolumn{7}{c}{Depth} &&
        \multicolumn{2}{c}{Edge} \\
    \cmidrule{3-9} \cmidrule{11-12}
    && Abs Rel$\downarrow$ & Sq Rel$\downarrow$ & RMSE$\downarrow$ & $\log_{10}\!\downarrow$ & $\delta_1\!\uparrow$ & $\delta_2\!\uparrow$ & $\delta_3\!\uparrow$ & & ORD$\downarrow$ & $\text{D}^\text{3}\text{R}\!\downarrow$ \\
    \midrule
    \multirow{6}{*}{DIML} & LeReS~\cite{leres}    
    & $0.101$ & $45.607$ & $325.191$ & $0.043$ & $0.902$ & $0.990$ & $0.998$ &&  
    $0.242$ & $0.284$ \\
    & Kim \textit{et al.}~\cite{layerrefine}  
    & $0.100$ & $45.554$ & $325.155$ & $0.042$ & $0.902$ & $0.990$ & $0.998$ && 
    $0.243$ & $0.301$ \\
    & Graph-GDSR~\cite{graph-GDSR}  
    & $0.101$ & $45.993$ & $326.320$ & $0.043$ & $0.901$ & $0.989$ & $0.998$ && 
    $0.243$ & $0.300$ \\
    & GBDF~\cite{gbdf}   
    & $0.100$ & $44.038$ & $\textbf{318.874}$ & $0.042$ & $0.906$ & $\textbf{0.991}$ & $0.998$ && 
    $0.239$ & $0.267$ \\
    & Boost~\cite{boostdepth}
    & $0.108$ & $50.923$ & $341.992$ & $0.046$ & $0.897$ & $0.987$ & $0.998$ && 
    $0.274$ & $0.438$ \\
    & Ours
    & $\textbf{0.098}$ & $\textbf{41.328}$ & $320.193$ & $\textbf{0.042}$ & $\textbf{0.926}$ & $0.990$ & $\textbf{0.998}$ &&  $\textbf{0.221}$ & $\textbf{0.230}$ \\
    \arrayrulecolor{black}\midrule
    
    \multirow{6}{*}{DIODE} & LeReS~\cite{leres}    
    & $0.105$ & $1.642$ & $9.856$ & $0.041$ & $0.892$ & $0.968$ & $\textbf{0.989}$ &&  
    $0.324$ & $0.685$ \\
    & Kim \textit{et al.}~\cite{layerrefine}  
    & $0.105$ & $1.654$ & $9.888$ & $0.044$ & $0.889$ & $0.964$ & $0.987$ && 
    $0.325$ & $0.713$\\
    & Graph-GDSR~\cite{graph-GDSR}  
    & $0.104$ & $1.626$ & $9.876$ & $0.044$ & $0.890$ & $0.967$ & $0.988$ && 
    $0.326$ & $0.690$\\
    & GBDF~\cite{gbdf}   
    & $0.105$ & $1.625$ & $9.770$ & $\textbf{0.041}$ & $0.894$ & $0.968$ & $0.990$ && 
    $0.322$ & $0.673$\\
    & Boost~\cite{boostdepth}
    & $0.105$ & $1.612$ & $9.879$ & $0.044$ & $0.892$ & $0.966$ & $0.987$ && 
    $0.343$ & $0.640$ \\
    & Ours
    & $\textbf{0.098}$ & $\textbf{1.529}$ & $\textbf{9.549}$ & $0.042$ & $\textbf{0.900}$ & $\textbf{0.968}$ & $0.988$ &&  $\textbf{0.293}$ & $\textbf{0.637}$ \\
    \arrayrulecolor{black}\bottomrule
    \end{tabular}
    }
\end{center}
\vspace{-5pt}
\caption{\textbf{Comparisons with previous refinement approaches on DIML and DIODE.}}
\label{tab:dimldiode_1s}
\end{table*}

\subsection{More Quantitative and Qualitative Results}
\noindent \textbf{Training Iterations of Self-distillation}
We investigate the iteration numbers of self-distillation in \reffig{}~\ref{fig:supp_iter}. The iteration number of zero indicates the model after the training of the first epoch only with ground truth for supervision, \textit{i.e.}, before self-distillation. Clearly, with the proposed self-distillation paradigm, both the depth accuracy and edge quality are improved until convergence.

\noindent \textbf{Formats of Pseudo-labels}
We compare the refined depth $D_S$ and the proposed depth edge representation $G_S$ as pseudo-labels. Using the accurate and meticulous depth $D_S$ could be a straightforward idea. However, with depth maps as the supervision, the model cannot precisely focus on improving edges and details. Thus, $G_S$ achieves stronger efficacy than $D_S$, proving the necessity of our designs.

\noindent \textbf{Quantitative Comparisons.} In the main paper, only $\delta_1$, REL, ORD, and $\text{D}^3\text{R}$ are reported. Here, we present the additional metrics of all the compared methods~\cite{layerrefine,graph-GDSR,gbdf,boostdepth,patchfusion} on Middlebury2021~\cite{middle}, DIML~\cite{diml}, and DIODE~\cite{diode} datasets in \reftab{}~\ref{tab:mid21_1s}, \reftab{}~\ref{tab:mid21_2s}, and \reftab{}~\ref{tab:dimldiode_1s}. Our method outperforms previous approaches on most evaluation metrics, showing the effectiveness of our \sx{} framework.

\begin{figure*}[]
\centering
\includegraphics[width=1\textwidth,trim=0 0 0 0,clip]{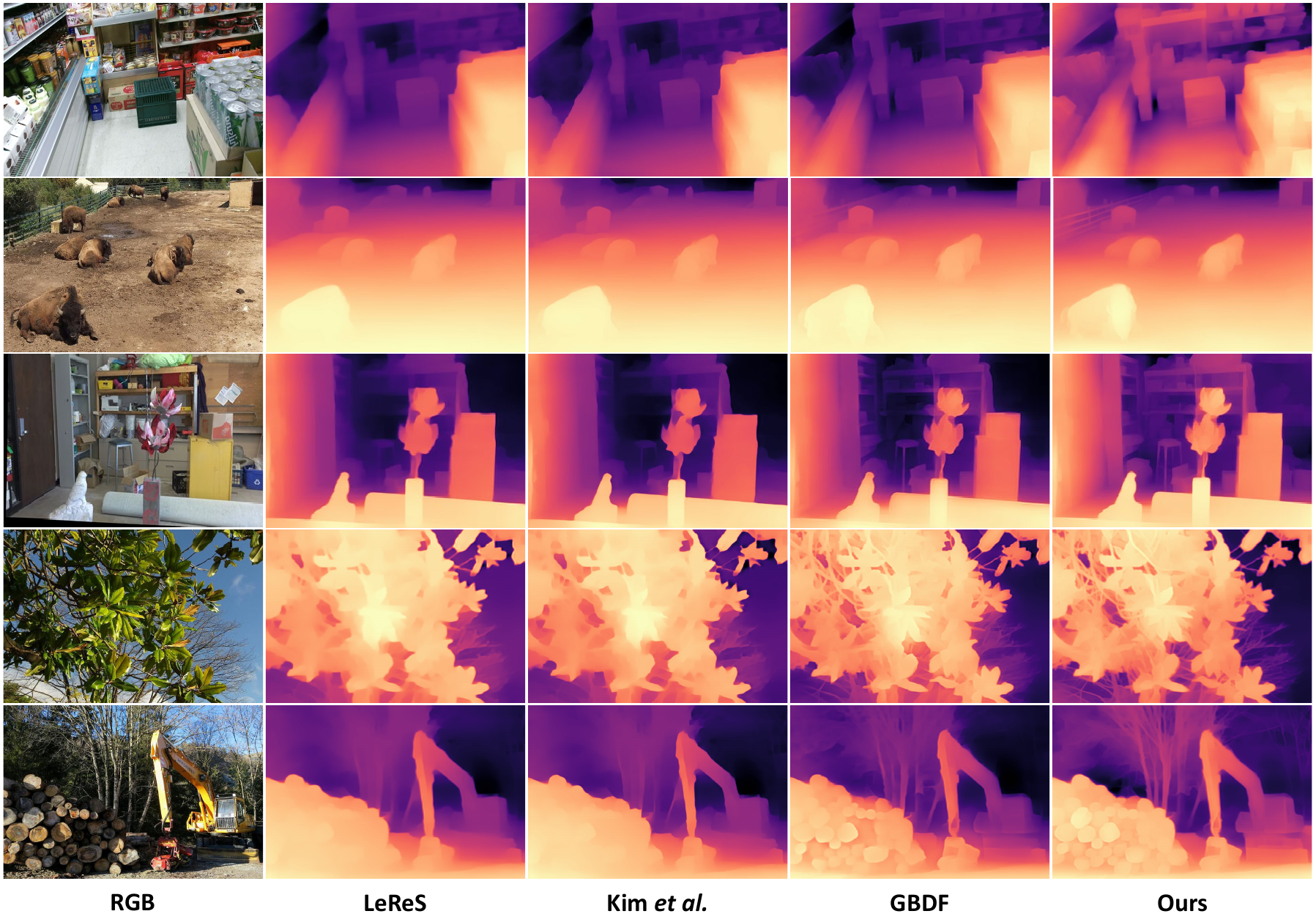}
\vspace{-10pt}
\caption{
\textbf{Qualitative comparisond with one-stage methods~\cite{layerrefine,gbdf} on various datasets~\cite{diml,kexian2020,middle}.} We adopt LeReS~\cite{leres} as the depth predictor. Better viewed when zoomed in.}
\label{fig:supp_1stage}
\end{figure*}

\begin{figure*}[htbp]
\vspace{-15pt}
\centering
\includegraphics[width=1\textwidth,trim=0 0 0 0,clip]{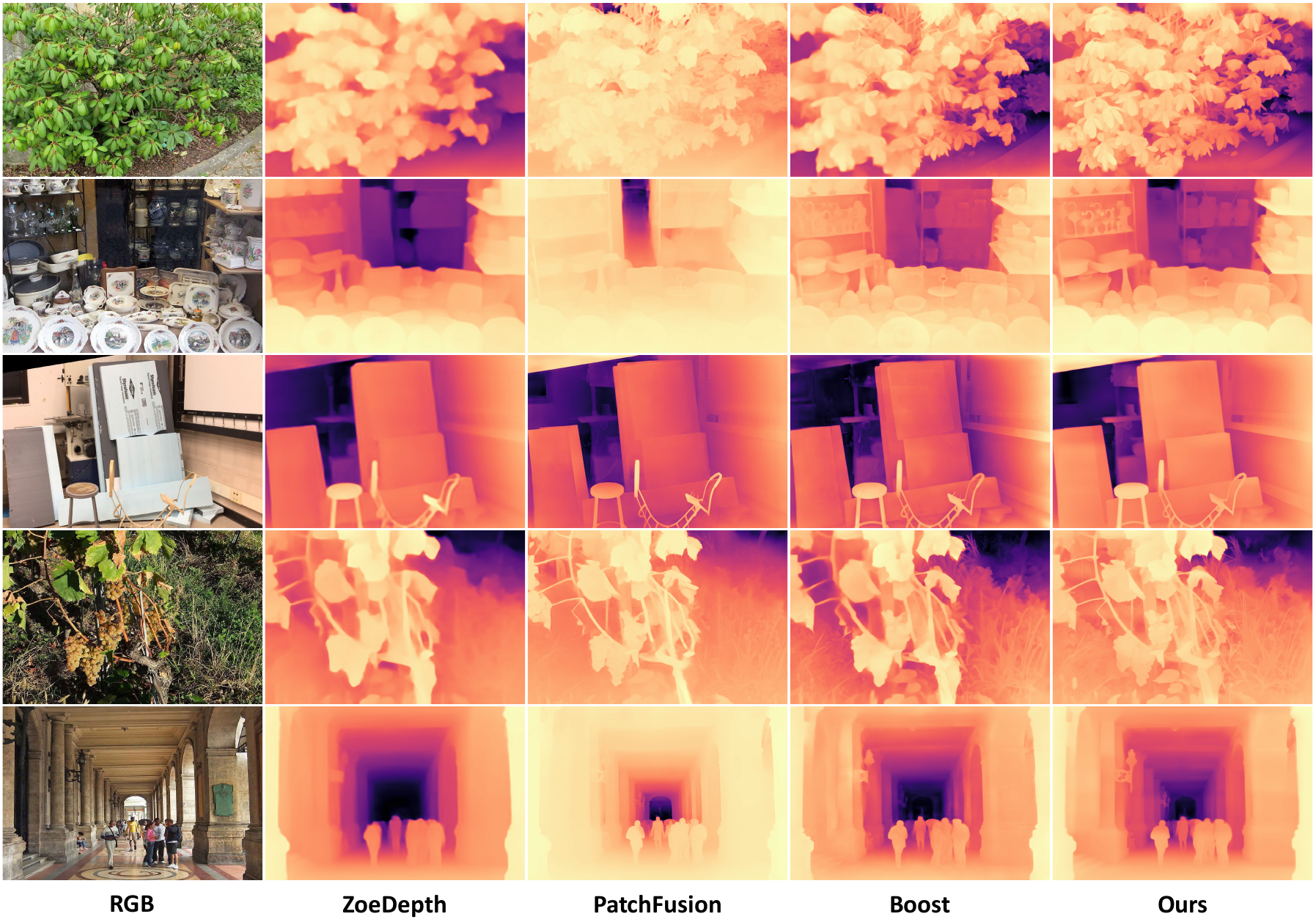}
\vspace{-10pt}
\caption{
\textbf{Qualitative comparisons with two-stage methods~\cite{patchfusion,boostdepth} on various datasets~\cite{diml,kexian2020,middle}.} We adopt Zoedepth~\cite{zoedepth} as the depth predictor. Better viewed when zoomed in.}
\label{fig:supp_2stage}
\end{figure*}

\noindent \textbf{Qualitative Comparison}
We provide more qualitative comparisons with one-stage~\cite{layerrefine,gbdf} and two-stage~\cite{patchfusion,boostdepth} methods in \reffig{}~\ref{fig:supp_1stage} and \reffig{}~\ref{fig:supp_2stage}. These visual results further demonstrate the excellent performance and generalization capability of \sx{} on diverse scenes~\cite{middle,diml,kexian2020}.

\end{document}